\documentclass[conference]{IEEEtran}
\IEEEoverridecommandlockouts
\usepackage{cite}
\usepackage{amsmath,amssymb,amsfonts}
\usepackage{graphicx}
\usepackage{textcomp}
\usepackage{xcolor}
\usepackage{diagbox}
\usepackage{siunitx}
\def\BibTeX{{\rm B\kern-.05em{\sc i\kern-.025em b}\kern-.08em
		T\kern-.1667em\lower.7ex\hbox{E}\kern-.125emX}}
	
\graphicspath{{figure/}}	

\usepackage{mathrsfs} 

\usepackage{amsthm}  
\usepackage{geometry}

\usepackage{algorithm,algorithmic}
\usepackage{bm}
\usepackage{url}
\usepackage[labelformat=simple]{subcaption}

\usepackage{multirow} 

\makeatletter
\renewcommand{\maketag@@@}[1]{\hbox{\m@th\normalsize\normalfont#1}}%
\makeatother    

\makeatletter

\newcommand{\Rmnum}[1]{\expandafter\@slowromancap\romannumeral #1@}
\makeatother

\newtheorem{definition}{Definition}

\newtheorem{theorem}{Theorem}
\newtheorem{lemma}{Lemma}

\def\thickhline{\noalign{\hrule height1pt}}

\begin{document}
	\newgeometry{top=54pt, bottom=54pt, left=54pt, right=54pt}
	\title{ \vspace{18pt} BESTAnP: \textbf{B}i-Step \textbf{E}fficient and \textbf{St}atistically Optimal Estimator for \textbf{A}coustic-\textbf{n}-\textbf{P}oint Problem
	}

\author{Wenliang Sheng$^{1*}$, Hongxu Zhao$^{2*}$, Lingpeng Chen$^{2*}$, Guangyang Zeng$^{2\dagger}$,\\ Yunling Shao$^{2}$, Yuze Hong$^{2}$, Chao Yang$^{1\dagger}$, Ziyang Hong$^{2}$, and Junfeng Wu$^{2}$
        \thanks{$^{1}$Key Laboratory of Smart Manufacturing in Energy Chemical Process, Ministry of Education, Dept. of Automation, East China University of Science and Technology, Shanghai, China.~$^{2}$The School of Data Science,
			Chinese University of Hong Kong, Shenzhen, Shenzhen, P. R. China.
			}%

            \thanks{$^*$ Equally contributed.  $^\dagger$ Corresponding authors. }
             \thanks{zengguangyang@cuhk.edu.cn, yangchao@ecust.edu.cn.}
		}%


	\maketitle
	
	\begin{abstract}
    We consider the acoustic-n-point (AnP) problem, which estimates the pose of a 2D forward-looking sonar (FLS) according to $n$ 3D-2D point correspondences.
    We explore the nature of the measured partial spherical coordinates and reveal their inherent relationships to translation and orientation. Based on this, we propose a bi-step efficient and statistically optimal AnP (BESTAnP) algorithm 
	that decouples the estimation of translation and orientation.
	Specifically, in the first step, the translation estimation is formulated as the range-based localization problem based on distance-only measurements. In the second step, the rotation is estimated via eigendecomposition based on azimuth-only measurements and the estimated translation. 
    BESTAnP is the first AnP algorithm that gives a closed-form solution for the full six-degree pose. In addition, we conduct bias elimination for BESTAnP such that it owns the statistical property of consistency. 
    Through simulation and real-world experiments, we demonstrate that compared with the state-of-the-art (SOTA) methods, BESTAnP is over ten times faster and features real-time capacity in resource-constrained platforms while exhibiting comparable accuracy. Moreover, for the first time, we embed BESTAnP into a sonar-based odometry which shows its effectiveness for trajectory estimation.


	\end{abstract}
	
	\section{Introduction}
		
\begin{figure}[htbp]
			\centering
			\includegraphics[width=8cm]{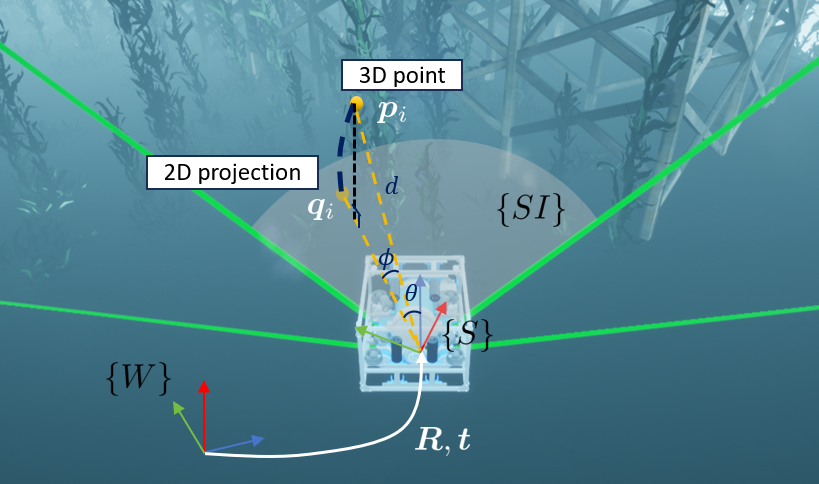}
			\caption{Illustration of the AnP problem. Given $n$ 3D points in the world frame and their corresponding 2D measurements in the sonar image, the pose of the FLS with respect to the world frame needs to be estimated.}
			\label{sonar model}
		\end{figure}

		The 2D forward-looking sonar (FLS) is frequently equipped in underwater robots and has been widely applied in the tasks of underwater navigation \cite{FRANCHI202014570}, object detection \cite{fuchs2018object}, and deep ocean exploring \cite{aal2000tectonic}. When equipping a 2D FLS, an autonomous underwater vehicle is able to measure the distance and azimuth of a 3D point.
		Similar to the perspective-n-point (PnP) problem for camera pose estimation, a basic problem, as illustrated in Fig.~\ref{sonar model}, arises for FLS pose estimation: given $n$ 3D points in the environment and their projections in the sonar image, how to estimate the pose (rotation and translation) of the sonar with respect to the environment?
		This problem is named as the acoustic-n-point (AnP) problem~\cite{wang2024acoustic}.

	The study for the AnP algorithm is of great value. In a pre-constructed map, it gives the absolute pose of an underwater robot, while in an unknown environment, it can be embedded into a sonar-based odometry or simultaneous localization and mapping (SLAM) system for autonomous navigation~\cite{wang2024acoustic,wang2020planar}.  
 However, there are difficulties in developing an AnP algorithm, such as the non-convexity of the problem and the lack of efficient feature extraction and association techniques for sonar images. As a result, the study of the AnP problem is still at its early stage. To the best of our knowledge, there is no AnP algorithm that can estimate the full six-degree pose of the 2D FLS with a closed form.

				
		In this letter, we propose a six-degree algorithm, named BESTAnP, with closed forms for both translation and rotation.
  The main contributions are summarized as follows:
  \begin{itemize}
\item \textbf{Six-degree closed-form solutions.} We reveal that distance measurements completely determine the translation, and azimuth measurements fully decode the rotation once the translation is known. Thanks to the decoupling of translation and rotation, we derive an efficient closed-form estimator. Experiments show that our algorithm is over ten times faster than the state-of-the-art (SOTA) methods and features real-time capacity in resource-constrained platforms.
\item \textbf{Consistent and statistically optimal properties.} We set out from the estimation theory and optimize the statistical property of our algorithm. By estimating the noise variances, bias elimination is conducted, yielding a consistent estimator that converges to the true pose as the measurement number increases. Moreover, we prove that a further GN iteration produces a statistically optimal estimator, with covariance approaching the Cramer-Rao lower bound (CRLB). 
\item \textbf{Comprehensive experimental evaluation.} Extensive synthetic and real-world experiments are conducted to evaluate the proposed AnP algorithm. Apart from absolute pose estimation, for the first time, we integrate the AnP algorithm with 3D triangulation to create sonar-based odometry for trajectory estimation. Results show that our BESTAnP algorithm matches SOTA accuracy while being significantly more efficient. All codes and data are open-sourced at \url{https://github.com/LIAS-CUHKSZ/BESTAnP}.
  \end{itemize}


	\section{Related Works}
		\subsection{PnP for Camera Pose Estimation}
		 Camera and FLS share
    similarity in the sense that both of them can only measure two dimensions of a 3D point. An FLS is able to measure the distance and azimuth of a point, while for a visual camera, the measurable information is the azimuth and elevation. The PnP problem refers to estimating the pose of a camera given $n$ 3D-2D point correspondences.  
		 PnP algorithms can be categorized into non-iterative and iterative methods. 
		 For non-iterative methods, the problem is usually solved by linear or polynomial solvers~\cite{lepetit2009ep,hesch2011direct,zheng2013revisiting,zhou2019efficient}.
		 For iterative methods, the PnP problem is formulated as a non-linear optimization problem, which is then solved using iterative methods such as the Gauss-Newton (GN) algorithm~\cite{pavlakos20176,chong2023introduction}.
   Non-iterative solutions are sometimes followed by local iterative refinement~\cite{urban2016mlpnp,zeng2023cpnp}.


		\subsection{AnP for Sonar Pose Estimation}
         
		The primary focus of underwater sonar geometry research lies in 3D motion estimation~\cite{negahdaripour20133,huang2016incremental,negahdaripour2008bundle}, 3D reconstruction~\cite{negahdaripour2018application,teixeira2019dense,rahman2019contour}, and SLAM~\cite{suresh2020active,westman2018feature,westman2019degeneracy}, with relatively less emphasis on the study of the AnP problem. However, like PnP, AnP is equally significant in the field of underwater navigation and constitutes a basic problem for sonar geometry study. For the AnP problem,  Wang \textit{et al.}~\cite{wang2020planar} utilized a linearized projection model similar to the weak perspective model of optical cameras to estimate the pose of acoustic cameras using coplanar feature points. However, this model does not accurately describe the process of acoustic projection. Wang \textit{et al.}~\cite{wang2024acoustic} considered the accurate acoustic projection model and proposed two methods to obtain closed-form solutions for five-degree pose estimation. 
  
        Note that the work~\cite{wang2020planar} confines 3D environment points in a plane. In addition, both~\cite{wang2020planar} and~\cite{wang2024acoustic} solve the third dimension of the translation using an iterative manner, which is not only computationally inefficient but also sensitive to the initial value. 

        \section{Problem Formulation}

        Throughout this paper, we adopt the notation convention that the superscript $(\cdot)^*$ denotes the true value or noise-free counterpart of a variable $(\cdot)$.

        \subsection{Ideal Measurement Model}
        The world frame $\{W\}$ is a global and fixed frame, and the sonar frame $\{S\}$ is a local coordinate system established on the sonar. The relative pose of $\{S\}$ with respect to $\{W\}$ can be described by a rotation matrix ${\bm R}^* \in {\rm SO}(3)$ and a translation vector ${\bm t}^* \in \mathbb R^3$, as shown in Fig.~\ref{sonar model}. Given $n$ points $^W \!\bm p_i, i=1,\ldots,n$ in the world frame, their coordinates in the sonar frame can be written as $^S \!\bm p_i = {\bm R}^{*\top}(^W \!\bm p_i - {\bm t}^*)$.

        Given the polar coordinates $[d_i^*~\theta_i^*~\phi_i^*]^\top$ in the sonar frame, where $d_i^*$, $\theta_i^*$, and $\phi_i^*$ are the distance, azimuth angle, and elevation angle, respectively, $^{S} \! \bm{p}_i$ can be represented as
        \begin{eqnarray} \label{polar_to_cartesian}
			^{S} \! \bm{p}_i =  \left[ \begin{array}{c} ^{S} \! x_i \\ ^{S} \! y_i \\ ^{S} \! z_i \end{array} \right] 
			= \left[ \begin{array}{c} d_i^* \cos \phi_i^* \cos \theta_i^* \\ d_i^* \cos \phi_i^* \sin\theta_i^* \\ d_i^* \sin \phi_i^* \end{array} \right]. 
		\end{eqnarray}
    As shown in Fig.~\ref{sonar model}, the projection of $^S \!\bm p_i$ onto the sonar image $\{SI\} $, where $\phi=0$, is 
        \begin{eqnarray} \label{projection_model}
			^{SI} \! \bm{q}_i =  \left[ \begin{array}{c} ^{SI} \! x_i \\ ^{SI} \! y_i \end{array} \right] 
			= \left[ \begin{array}{c} d_i^*	\cos\theta_i^* \\ d_i^* \sin\theta_i^* \end{array} \right]. \label{1}
      \end{eqnarray}
     
        The projection model~\eqref{projection_model} shows that sonar can measure the distance and azimuth angle of a target, while the elevation angle is unobservable. Note that $d_i^* = \Vert^{S} \! \bm p_i\Vert$ and $\theta_i^* = \tan^{-1}(^S\! y_i/^S \! x_i)$. According to $^S \!\bm p_i = {\bm R}^{*\top}(^W \!\bm p_i - {\bm t}^*)$ and the norm-preserving property of a rotation matrix, the ideal measurement model is 
        \begin{equation}
                d_i^* = \Vert ^W \! \bm p_i - \bm t^* \Vert,~
            \theta_i^* = \tan^{-1}\left(\frac{\bm r_2^*(^W \! \bm p_i - \bm t^*)}{\bm r_1^*(^W \! \bm p_i - \bm t^*)}\right), 
            \label{theta_true}
        \end{equation}
        where $\bm r_i^*$ is the $i$-th row of $\bm R^{*\top}$.


    \subsection{Problem Formulation}
   We has introduced the noise-free sonar measurement model. In practice, sonar measurements contain noise. We consider the following measurement model:
    \begin{equation}
             d_i = \Vert ^{W} \! \bm{p}_i - {\bm t^*} \Vert + \epsilon_{d_i}, ~ 
         \tan\theta_i = \frac{\bm r_2^*(^W \! \bm p_i - \bm t^*)}{\bm r_1^*(^W \! \bm p_i - \bm t^*)} + \epsilon_{\theta_i}, 
        \label{measurement model}
    \end{equation}
    where $\epsilon_{d_i} \sim \mathcal N(0,\sigma_d^2)$ and $\epsilon_{\theta_i} \sim \mathcal N(0,\sigma_{\theta}^2)$ are independent Gaussian noises.
Adding noise to $\tan \theta_i^*$ instead of $\theta_i^*$ is mainly for the convenience of theoretical derivation in the next section. This noise-adding mechanism is also adopted in literature involving angle measurements~\cite{zeng2023cpnp,rahman2022svin2,lepetit2009ep}. It is also noteworthy that even if the noise is directly added to $\theta_i^*$, when the azimuth angle aperture is not large and the noise intensity is small, as most FLS are,~\eqref{measurement model} is still a good first-order approximation. The rationality for this noise-adding modeling will also be validated in the experiments in Section~\ref{section:5}.

    Define $f_{d_i}({\bm t}) \!=\! d_i - \Vert^W \! \bm p_i - \bm t\Vert$ and $f_{\theta_i}({\bm R},{\bm t}) \!=\! \tan\theta_i-\frac{\bm e_2^\top\bm R^\top({^W \! \bm p_i} - \bm t)}{\bm e_1^\top\bm R^\top(^W \! \bm p_i - \bm t)}$, where $\bm e_i$ is the unit vector whose $i$-th element is 1. Given the world coordinates of $n$ 3D points and their corresponding measurements $d_i$ and $\theta_i$, we formulate the following maximum likelihood (ML) problem to estimate the sonar pose:
        \begin{equation} \label{optimization_problem}
       \begin{split}
            \underset{\bm{R}, \bm{t}}{\operatorname{minimize}} ~& \frac{1}{n} \sum_{i=1}^n \left({f_{d_i}({\bm t})^2 \over \sigma_d^2} + {f_{\theta_i}({\bm R},{\bm t})^2 \over \sigma_\theta^2} \right) \\
            ~~~\text { subject to } & \bm {R} \in \mathrm{SO}(3),~{\bm t} \in \mathbb R^3.
            \end{split}
        \end{equation}
        
    A global solution, denoted as $\hat{\bm R}^\mathrm{ML}$ and $\hat{\bm t}^\mathrm{ML}$, to~\eqref{optimization_problem} is called the ML estimate. The ML problem~\eqref{optimization_problem} is non-convex, making it difficult to solve. In what follows, we decouple the estimation for $\bm R$ and $\bm t$ into two steps and design consistent estimators for them respectively.

    \section{BESTAnP algorithm}
        In this section, we explore the nature presented by the spherical coordinates of the two measured dimensions, the distance $d_i$ and the azimuth angle $\theta_i$, and show their inherent relations to the translation and rotation. First, we find that the measured distances fully determine the location of the origin of the sonar frame. Second, when the sonar frame's location is obtained, the azimuth angles fully determine the rotation of the sonar frame. 
        We propose the BESTAnP algorithm, in which we sequentially design consistent estimators for translation and rotation, respectively. A GN iteration is used to further improve the accuracy of the consistent estimators.
        \subsection{Consistent Estimator of the Translation Vector}
    
         When a point  $^W \! \bm p_i$ is measured by sonar with a distance $d_i$, the origin of the sonar should be on the surface of a sphere centered at $^W \! \bm p_i$ with radius $d_i$. With multiple $^W \! \bm p_i$'s and $d_i$'s,  the spheres so defined intersect at the origin.
         However, due to measurement noise, 
   exact intersection does not happen.      
        Estimating the position of the sonar origin in the world frame, i.e., the translation vector $\bm t^*$, is essentially a range-based localization problem~\cite{xu2011source,nguyen2019distance,9855392}. We estimate $\bm t^*$ as well as the variance $\sigma_d^2$ of distance measurement noises using the consistent estimation method presented in~\cite{9855392}. Specifically, the following bias-eliminated linear least-squares problem is constructed:
        \begin{equation} \label{BELLS}
            \begin{aligned}
            &\begin{aligned}
            \underset{\bm {x} \in \mathbb{R}^{4}}{\operatorname{minimize}} ~&  \frac{1}{n}\Vert \bm A \bm x - \bm b\Vert^2,
            \end{aligned}
            \end{aligned}
        \end{equation}
        where 
        \begin{eqnarray*}
            \bm A=\left[ \begin{array}{cc} 
                -2 ~^W\!\bm p_1^\top & 1\\
                \vdots& \vdots\\
                -2 ~ ^W\!\bm p_n^\top & 1
               \end{array} \right], 
            ~~\bm b =\left[\begin{array}{c}
               d_1^2-\Vert ^W\!\bm p_1 \Vert^2 \\
               \vdots\\
                d_n^2-\Vert ^W\!\bm p_n \Vert^2	   
           \end{array}\right].
		\end{eqnarray*}
        
        When the 3D points $^W\bm p_i$'s are spatially non-coplanar, the optimal solution to \eqref{BELLS} is $\hat{\bm x}^\mathrm{BE} =(\bm A^\top \bm A)^{-1}\bm A^\top\bm b$. Then, the estimator for $\bm t^*$ and $\sigma_d^2$ is 
\begin{equation}
                 \hat{\bm t}^\mathrm{BE} = [\hat{\bm x}^{\rm BE}]_{1:3},~
            \hat \sigma_d^2 = [\hat{\bm x}^{\rm BE}]_{4}-\|[\hat{\bm x}^{\rm BE}]_{1:3}\|^2, 
    \label{consistent_estimate_t}
        \end{equation}
        where $[\hat{\bm x}^{\rm BE}]_{1:3}$ denotes the vector containing the first three elements of $\hat{\bm x}^{\rm BE}$, and $[\hat{\bm x}^{\rm BE}]_{4}$  the fourth element. It is shown in~\cite{9855392} that the estimator~\eqref{consistent_estimate_t} is $\sqrt{n}$-consistent. The definition of a $\sqrt{n}$-consistent estimator is given as follows:
        \begin{definition}[Big $O_p$]
            The estimator $\hat {\bm \gamma}$ is called a $\sqrt{n}$-consistent estimator of ${\bm \gamma}^*$ if $\hat {\bm \gamma}-{\bm \gamma}^*=O_p(1/\sqrt{n})$, i.e., for any $\varepsilon >0$, there exists a finite $M$ and a finite $N$ such that for any $n>N$, $\mathbb{P} (\|\sqrt{n}(\hat {\bm\gamma} -{\bm\gamma}^*)\|>M )<\varepsilon$.
        \end{definition}
        ``$\sqrt{n}$-consistent'' includes two implications: The estimator is consistent, i.e., it converges to the true value as $n$ increases; The convergence is as fast as $1/\sqrt{n}$. \begin{theorem}{(\cite{9855392}, Theorem 4)}\label{theorem_t}
            The estimator~\eqref{consistent_estimate_t} is $\sqrt{n}$-consistent for both $\bm t^*$ and $\sigma_d^2$.
        \end{theorem}

    \subsection{Consistent Estimator of the Rotation Matrix}
        First, we consider the case that the azimuth angle measurements are noise-free and the true translation ${\bm t}^*$ is given. According to \eqref{theta_true}, 
        \begin{align}
            \tan\theta_i^*\bm r_1^*(^W \! \bm p_i - \bm t^*) - \bm r_2^*(^W \! \bm p_i - \bm t^*) = 0.
        \end{align}
        Then, we can calculate ${\bm r}^*=[\bm r_1^*~\bm r_2^*]^\top$ by solving the following optimization problem: 
        
        \begin{equation}  \label{optimizaiton_R_noisefree}
        \begin{split}
            \underset{\bm{r} \in \mathbb R^6}{\operatorname{minimize}} ~& \frac{1}{n}\Vert{\bm {B^*}} {\bm {r}}\Vert^2 \\
            ~~~\text { subject to } & \Vert{\bm {r}}\Vert^2 = 2,
            \end{split}
        \end{equation}
        where
        		\begin{eqnarray*}
			\bm B^* \!=\! \left[\begin{array}{cc}
				\tan\theta_1^*( ^W \! \bm p_1 - {\bm t}^*)^\top &  ({\bm t}^* -^W \! \bm p_1)^\top \\ 
				\vdots & \vdots \\
				\tan\theta_n^*( ^W \! \bm p_n - {\bm t}^*)^\top &  ({\bm t}^* -^W \! \bm p_n)^\top
			\end{array}\right].
		\end{eqnarray*}
        
        Note that ${\bm B}^*{\bm r}^*=0$, thus $\text{rank}(\bm B^*) \leq 5$. When the 3D points are in generic positions, we have $\text{rank}(\bm B^*) = 5$. This guarantees that the optimal solution $\hat {\bm r}^*$ to~\eqref{optimizaiton_R_noisefree} is equal to ${\bm r}^*$ up to a sign.
        Denote the unit eigenvector corresponding to the smallest eigenvalue of the matrix $\bm Q^*={\bm B^{*\top}\bm B^*}/n$ as $\bm v_{\lambda_{\rm min}}^*$. Then $\hat {\bm r}^*$ can be calculated as ${\hat{\bm {r}}^*} = \pm \sqrt{2}~{\bm v}_{\lambda_{\rm min}}^*$~\cite{zeng2024optimal}.
        However, ${\bm Q}^*$ comprises noise-free azimuth angle measurements $\tan \theta_i^*$ and the true translation ${\bm t}^*$, which are both unavailable in practice. In what follows, we will construct a matrix ${\bm Q}^{\rm BE}$ from available noisy measurements $\tan \theta_i$ and estimate $\hat{\bm t}^\mathrm{BE}$ and prove that it converges to ${\bm Q}^*$ as $n$ increases. Then, we apply eigendecomposition for ${\bm Q}^{\rm BE}$ to construct a consistent estimator of ${\bm r}^*$.
        Before that, we show how to estimate the noise variance of azimuth angle measurements. Let ${\bm Q}={\bm B}^\top {\bm B}/n$, where 
        \begin{equation*}
            \bm B \!=\! \left[\begin{array}{cc}
				\tan\theta_1( ^W \! \bm p_1 - \hat{\bm t}^\mathrm{BE})^\top &  (\hat{\bm t}^\mathrm{BE} -^W \! \bm p_1)^\top \\ 
				\vdots & \vdots \\
				\tan\theta_n( ^W \! \bm p_n - \hat{\bm t}^\mathrm{BE})^\top &  (\hat{\bm t}^\mathrm{BE} -^W \! \bm p_n)^\top
			\end{array}\right].
        \end{equation*}
 Set
        \begin{align*}
            \bm{S} = \left[\begin{array}{cc}
				{1\over n}\sum_{i=1}^{n}\bm (^W \! \bm p_i - \hat{\bm t}^\mathrm{BE})(^W \! \bm p_i - \hat{\bm t}^\mathrm{BE})^\top &  \bm 0_{3 \times 3} \\
				\bm 0_{3 \times 3} & \bm 0_{3 \times 3}
			\end{array}\right].
        \end{align*}
         The estimate of $\sigma_{\theta}^2$ is obtained as 
    \begin{equation} \label{angle_noise_estimator}
    \hat\sigma^2_{\theta}=1/\lambda_{\rm max}({\bf Q}^{-1} {\bf S}),
    \end{equation}
    where $\lambda_{\rm max}(\cdot)$ denotes the largest eigenvalue of a matrix.
    
    \begin{lemma} \label{angle_noise_estimation}
    The estimator~\eqref{angle_noise_estimator} is 
$\sqrt{n}$-consistent for $\sigma_\theta^2$.
\end{lemma}
The proof of Lemma~\ref{angle_noise_estimation} is presented in Appendix~\ref{proof_angle_noise}.

        To eliminate the bias between ${\bm Q}$ and ${\bm Q}^*$, we design the correction matrix ${\bm C}=\hat \sigma_{\theta}^2 {\bm S}$.
          Let $\bm Q^\mathrm{BE} = \bm Q -\bm C$ and denote a unit eigenvector corresponding to the smallest eigenvalue of $\bm Q^\mathrm{BE}$ as ${\bm v}_{\lambda_{\rm min}}^\mathrm{BE}$.
        Our bias-eliminated estimator (up to a sign) for $\bm r^*$ is given by
        \begin{align} \label{consistent_estimate_r}
            \hat{\bm r}^\mathrm{BE} = \pm \sqrt{2}~{\bm v}_{\lambda_{\rm min}}^\mathrm{BE}.
        \end{align}
      Theoretically, the projected azimuth angle $\hat{\theta}_i$ based on the estimated pose should match the measurement $\theta_i$. Hence, we determine the sign of $\hat{\bm r}^\mathrm{BE}$ by checking whether $\cos\hat{\theta}_i$ and $\cos\theta_i$ have the same sign. With abuse of notation, we still denote the sign-corrected result as $\hat{\bm r}^\mathrm{BE}$.

         \begin{theorem}\label{theorem_Q}
            The matrix $\bm Q^\mathrm{BE}$ is a $\sqrt{n}$-consistent estimator of $\bm Q^*$. Moreover, $\hat{\bm r}^\mathrm{BE}$ is a $\sqrt{n}$-consistent estimator of $\bm r^*$.
        \end{theorem}

        The proof of Theorem~\ref{theorem_Q} is presented in Appendix~\ref{appendix_consistent_R}.
        
        Let $\hat{\bm {r}}_1^\mathrm{BE}$ and $\hat{\bm {r}}_2^\mathrm{BE}$ consist of the first and last three elements of $\hat{\bm {r}}^{\mathrm{BE}\top}$ and $\hat{\bm {r}}_3^\mathrm{BE}=\hat{\bm {r}}_1^\mathrm{BE}\times \hat{\bm {r}}_2^\mathrm{BE}$. Then, we obtain the rotation estimate $\hat{\bm R}^{BE} = [\bm{r}_1^{\mathrm{BE}\top}~\bm{r}_2^{\mathrm{BE}\top}~\bm{r}_3^{\mathrm{BE}\top} ]$. As $\hat{\bm R}^{BE}$ may not satisfy the ${\rm SO}(3)$ constraint, we need to project it onto the ${\rm SO}(3)$ group. Denote the SVD of $\hat{\bm R}^\mathrm{BE}$ by $\hat{\bm R}^\mathrm{BE}=\bm U_R \bm \Sigma_R \bm V_R^\top$, then the projection is given by $\bm U_R {\rm diag}([1,1,\det(\bm U_R\bm V_R^{\top})]) \bm V_R^\top$~\cite{zeng2023cpnp}. With a little abuse of notation, we still denote the final projection result as $\hat{\bm R}^\mathrm{BE}$. Since $\hat{\bm r}^\mathrm{BE}$ is $\sqrt{n}$-consistent, and the cross product and SVD are both continuous mappings, the $\sqrt{n}$-consistent property can be preserved~\cite{zeng2024optimal} and thus $\hat{\bm R}^\mathrm{BE}$ is also $\sqrt{n}$-consistent.

        \subsection{Gauss-Newton Refinement}
        We have obtained consistent estimators $\hat{\bm R}^\mathrm{BE}$ and $\hat{\bm t}^\mathrm{BE}$ which converge to the true pose $\bm R^*$ and $\bm t^*$ as $n$ increases. However, they do not have the minimum covariance. 
        We know from estimation theory that an ML estimator is asymptotically optimal under some mild regularity conditions, that is, its covariance converges to the CRLB~\cite{9855392}. 
        In this subsection, we will come back to the ML problem~\eqref{optimization_problem}, taking the consistent estimator as the initial value and applying the GN algorithm to seek the ML estimate. 
        Thanks to the $\sqrt{n}$ consistency of the initial value $\hat{\bm R}^\mathrm{BE}$ and $\hat{\bm t}^\mathrm{BE}$, an interesting result is that only a single GN iteration is enough to realize the same asymptotic property as the ML estimator, which is formally stated in Theorem~\ref{theorem_two_step}. Before that, we give the definition of small $o_p$.
         \begin{definition}[Small $o_p$]
            The notation $\hat {\bm \gamma}-{\bm \gamma}=o_p(1/\sqrt{n})$ means that $\sqrt{n}(\hat {\bm \gamma}-{\bm \gamma})$ converges to 0 in probability, i.e., for any $\varepsilon >0$, $\lim_{n \rightarrow \infty}\mathbb{P} (\|\sqrt{n}(\hat {\bm\gamma} -{\bm\gamma})\|>\varepsilon )=0$.
        \end{definition}
        \begin{theorem} \label{theorem_two_step}
	Denote the single GN iteration of $\hat{\bm R}^\mathrm{BE}$ and $\hat{\bm t}^\mathrm{BE}$ by $\hat{\bm R}^\mathrm{GN}$ and $\hat{\bm t}^\mathrm{GN}$. Then we have 
	\begin{equation*}
		\hat{\bm R}^\mathrm{GN}-\hat{\bm R}^\mathrm{ML}=o_p(1/\sqrt{n}),~\hat{\bm t}^\mathrm{GN}-\hat{\bm t}^\mathrm{ML}=o_p(1/\sqrt{n}).
	\end{equation*}
\end{theorem}
For the proof of Theorem~\ref{theorem_two_step}, please refer to the proof of Theorem 7 in~\cite{zeng2024optimal}. Theorem~\ref{theorem_two_step} implies that $\hat{\bm R}^\mathrm{GN}$ and $\hat{\bm t}^\mathrm{GN}$ have the same asymptotic property that $\hat{\bm R}^\mathrm{ML}$ and $\hat{\bm t}^\mathrm{ML}$ possess. In other words, $\hat{\bm R}^\mathrm{GN}$ and $\hat{\bm t}^\mathrm{GN}$ are statistically optimal and can asymptotically reach the CRLB as $n$ increases.  
        Regarding the ML problem~\eqref{optimization_problem}, the form of the GN iteration is presented in Appendix~\ref{formula_GN}. 

 \begin{figure}[t]
    \centering
   \begin{minipage}{0.49\linewidth}
       \includegraphics[width=\linewidth]{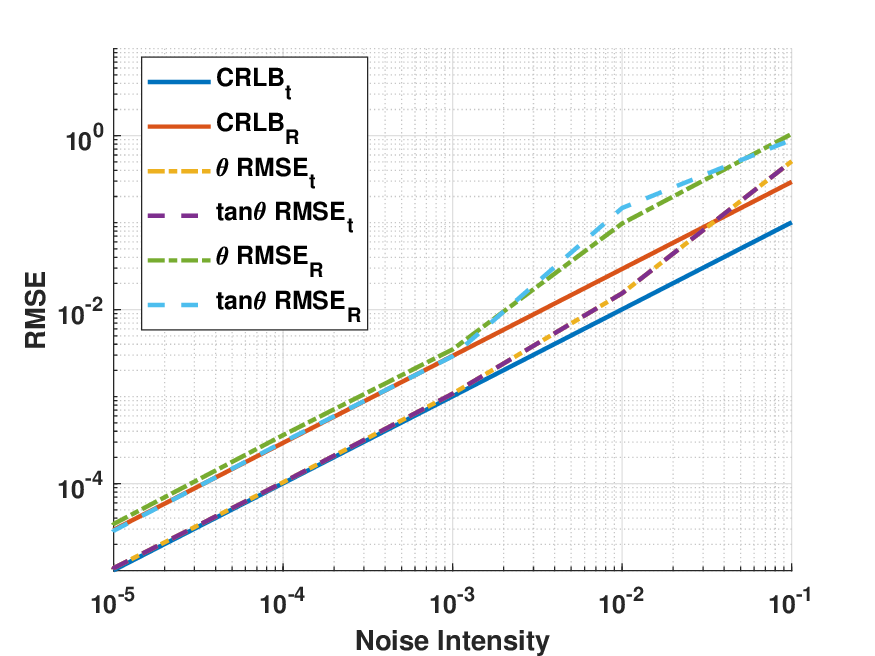}
            \caption{Effect of different noise-adding schemes.}
            \label{Noise_injection_way}
   \end{minipage}
    \begin{minipage}{0.49\linewidth}
       \includegraphics[width=\linewidth]{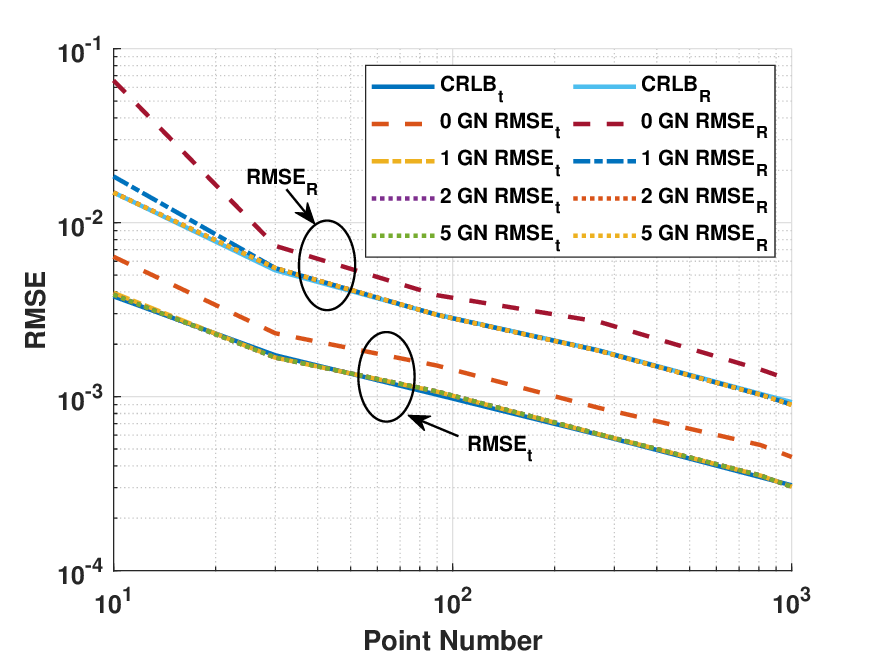}
            \caption{Effect of different GN iteration numbers.}
            \label{GN_compare}
   \end{minipage}
\end{figure}

\begin{figure}[t]
    \centering
   \begin{subfigure}{0.49\linewidth}
       \includegraphics[width=\linewidth]{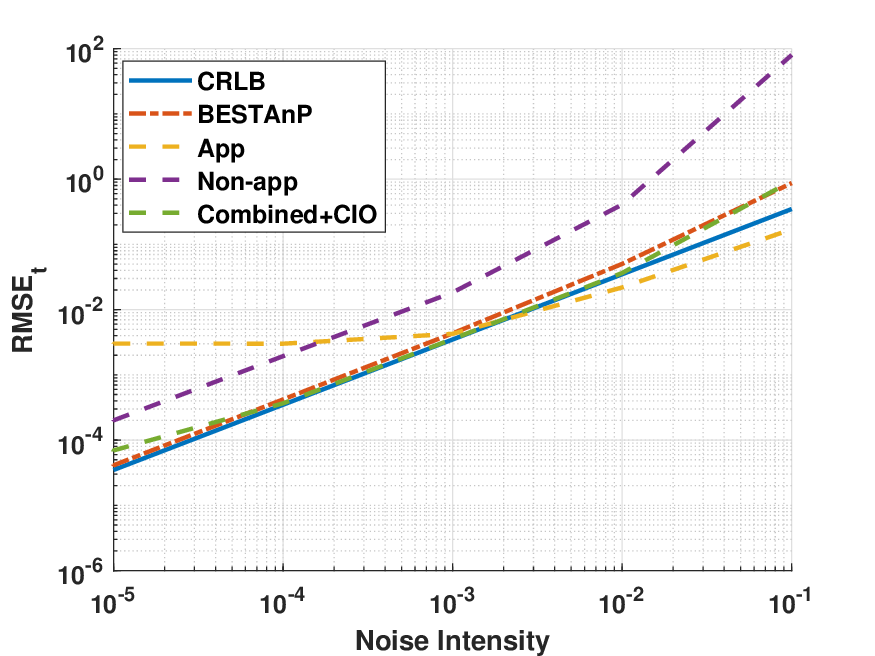}
   \end{subfigure}
    \begin{subfigure}{0.49\linewidth}
       \includegraphics[width=\linewidth]{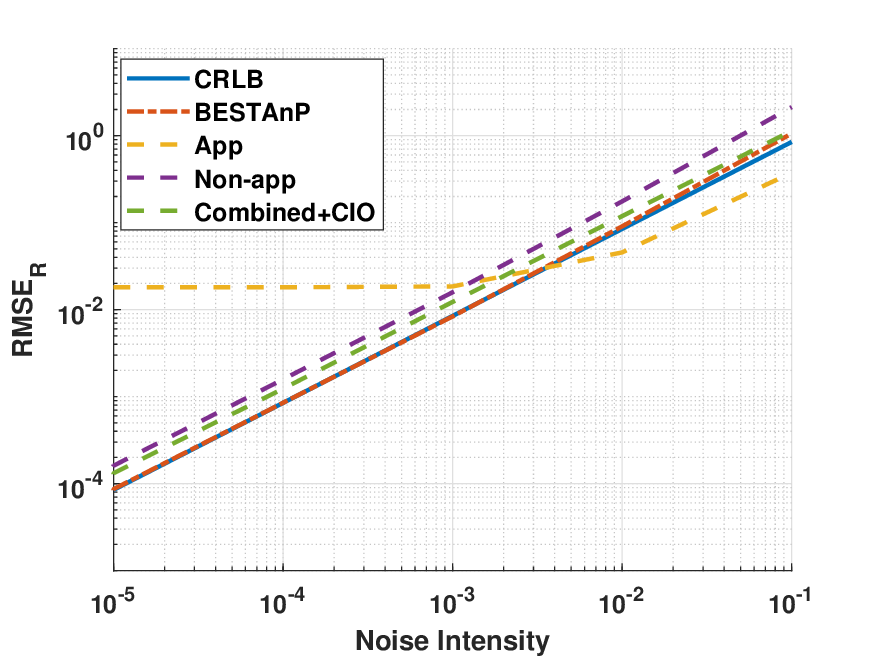}
   \end{subfigure}
   \caption{RMSE comparison under varied noise intensities.}
            \label{RMSE_varied_noise}
\end{figure}

        At the end of this section, we summarize our proposed BESTAnP algorithm in Algorithm~\ref{algorithm_summary}.

    \section{Experiments on synthetic data} \label{section:5}
    In this section, we evaluate the proposed algorithm by simulation. We not only test the accuracy of absolute pose estimation in a known environment with precise 3D coordinates but also embed the AnP algorithm into a sonar-based odometry for trajectory estimation in an unknown environment. 
    The proposed AnP algorithm is denoted as BESTAnP. 
    The compared methods include App (the approximated algorithm in~\cite{wang2024acoustic}), Non-app (the non-approximated algorithm in~\cite{wang2024acoustic}), and Combined+CIO (the algorithm in~\cite{wang2024acoustic} that uses a constrained iterative optimization method to refine the initial estimate selected from the App and Non-app solutions). 

\begin{figure}[t]
    \centering
   \begin{subfigure}{0.49\linewidth}
       \includegraphics[width=\linewidth]{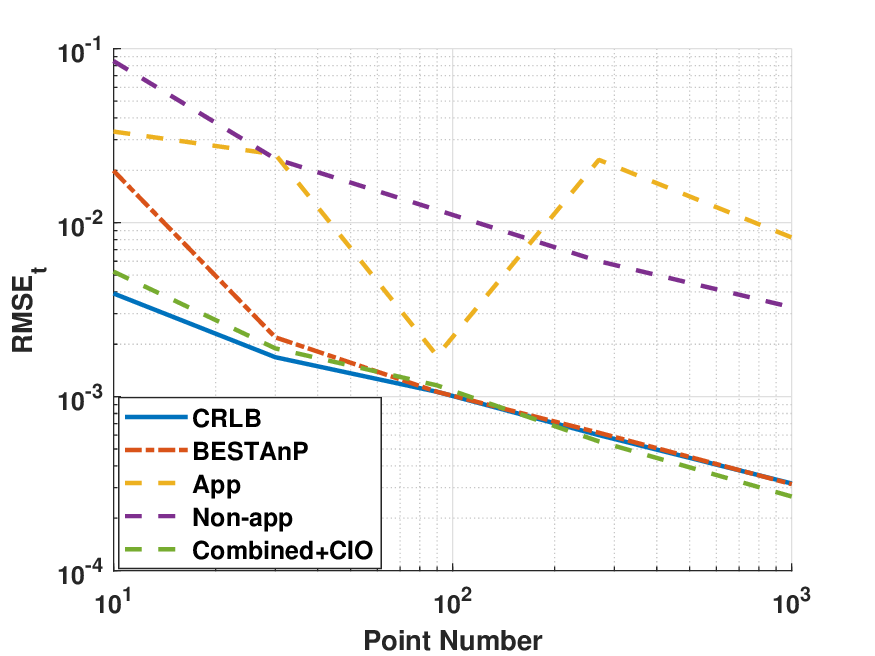}
   \end{subfigure}
    \begin{subfigure}{0.49\linewidth}
       \includegraphics[width=\linewidth]{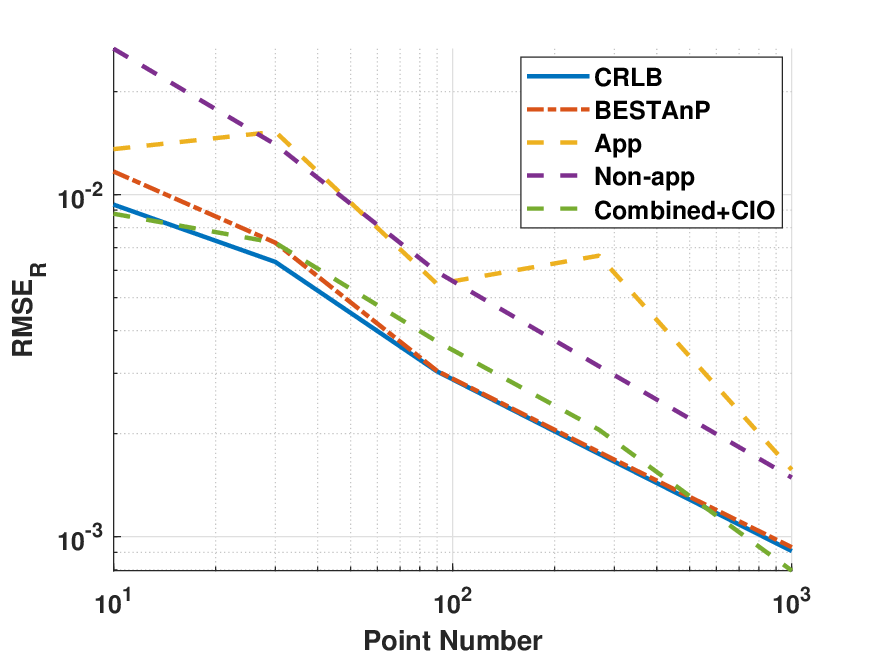}
   \end{subfigure}
   \caption{RMSE comparison under varied point numbers.}
            \label{RMSE_varied_pt_num}
\end{figure}

\begin{figure}[t]
    \centering
   \begin{subfigure}{0.49\linewidth}
       \includegraphics[width=\linewidth]{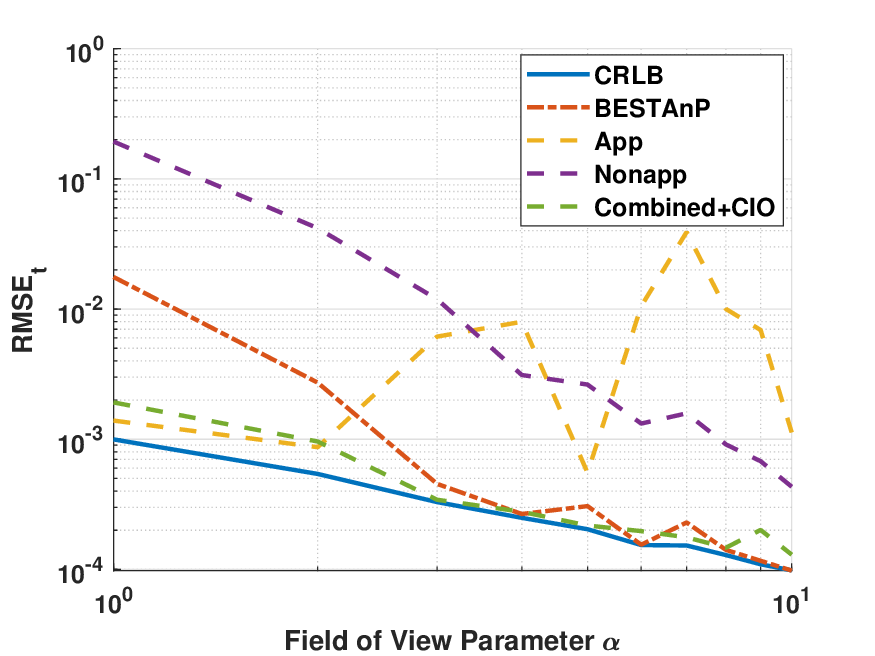}
   \end{subfigure}
    \begin{subfigure}{0.49\linewidth}
       \includegraphics[width=\linewidth]{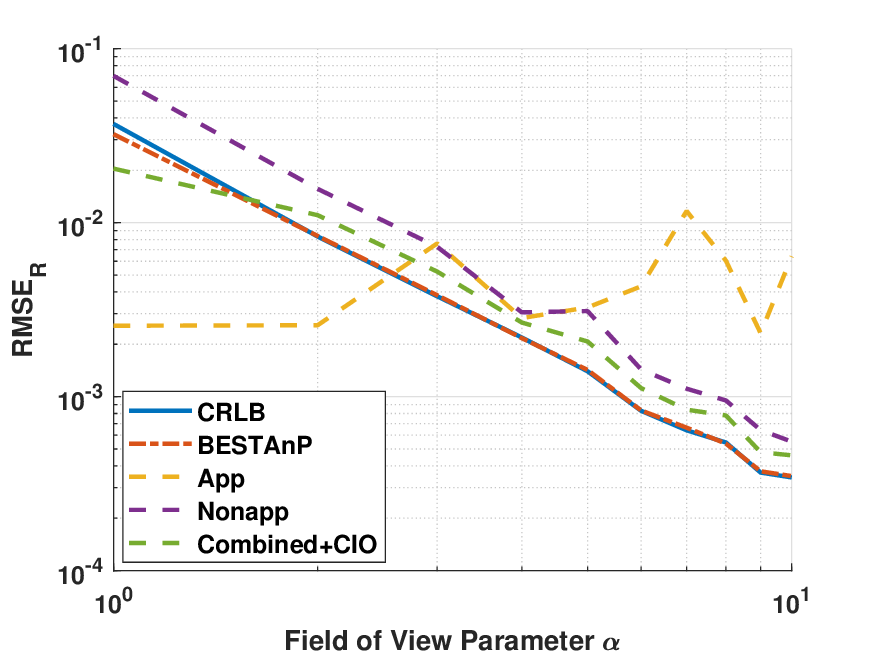}
   \end{subfigure}
   \caption{RMSE comparison under varied fields of view.}
            \label{RMSE_varied_pt_dis}
\end{figure}

\begin{algorithm}[t]
			\renewcommand{\algorithmicrequire}{\textbf{Input:}}
			\renewcommand{\algorithmicensure}{\textbf{Output:}}
			\caption{BESTAnP Algorithm}
			\label{algorithm_summary}
			\begin{algorithmic}[1] 
				\REQUIRE  3D world coordinates $^W \! \bm p_i$, distance measurements $d_i$, and azimuth angle measurements $\tan \theta_i$.  
				\ENSURE sonar pose estimate $\hat{\bm{R}}^{\mathrm{GN}}$,  $\hat{\bm{t}}^{\mathrm{GN}}$. 
				\STATE Calculate the translation estimate $\hat{\bm{t}}^{\mathrm{BE}}$ using \eqref{consistent_estimate_t}.
				\STATE Calculate the corrected matrix $\bm {Q}^{\mathrm{BE}}$ in Theorem~\ref{theorem_Q}.
				\STATE Perform eigendecomposition on $\bm {Q}^{\mathrm{BE}}$ and obtain  $\hat{\bm{r}}^{\mathrm{BE}}$ in \eqref{consistent_estimate_r}.
                \STATE Determine the sign of $\hat{\bm{r}}^{\mathrm{BE}}$ by checking the sign of $\cos \theta_1$ and the reprojected $\cos\hat{\theta}_1$. 
				\STATE Calculate $\hat{\bm{r}}_3^{\mathrm{BE}} = \hat{\bm {r}}_1^{\mathrm{BE}}\times\hat{\bm {r}}_2^{\mathrm{BE}}$ and obtain $\hat{\bm R}^{BE} = [\bm{r}_1^{\mathrm{BE}\top}~\bm{r}_2^{\mathrm{BE}\top}~\bm{r}_3^{\mathrm{BE}\top} ]$.
                \STATE Project $\hat{\bm{R}}^{\mathrm{BE}}$ onto ${\rm SO}(3)$.
                \STATE Calculate $\hat{\bm{R}}^{\mathrm{GN}}$ and $\hat{\bm{t}}^{\mathrm{GN}}$ using a single GN iteration.
			\end{algorithmic}
		\end{algorithm}
    
        \subsection{Absolute Pose Estimation with Known 3D Points}
        \label{AnP Test on Synthetic Data}
       The theoretical results developed in the previous section will be verified here. 
        We assume that the forward-looking sonar can detect targets within a distance of 6 meters, an azimuth angle range of [-30°, 30°], and an elevation angle range of [-10°, 10°]. Within these ranges, we first generate 3D points $^S\!\bm p_i$ in the sonar frame. Given the rotation matrix $\bm R^*$ and translation vector $\bm t^*$ of the sonar, the coordinates $^W \! \bm p_i$ in the world frame are obtained. According to the measurement model~\eqref{measurement model}, noises are added to $d_i^*$ and $\tan \theta_i^*$. The sonar's pose is then estimated using noisy measurements $d_i$ and $\tan \theta_i$ along with their corresponding world coordinates $^W\! \bm p_i$.

        Each experiment involves $T=1000$ Monte Carlo tests. We use the following root mean squared error (RMSE) to evaluate the estimation performance of each algorithm:
        \begin{equation*}
            \mathrm{RMSE}_{\bm{R}} \!= \!\sqrt{\frac{1}{T} \sum_{k=1}^T\|\hat{\bm{s}}_k\|^2}, ~
            \mathrm{RMSE}_{\bm{t}} \! =\!\sqrt{\frac{1}{T} \sum_{k=1}^T\|\hat{\bm{t}}_k-\bm{t}^*\|^2},
        \end{equation*}
        where $\hat{\bm{s}}_k = {\rm Log}(\bm{R}^{*\top}\hat{\bm{R}}_{k})$, 
         $\hat{\bm{R}}_{k}$ and $\hat{\bm{t}}_k$ are the
         estimates in the $k$-th Monte Carlo test, and ${\rm Log: {\rm SO}(3) \rightarrow \{{\bm s \in \mathbb R^3 \mid \|{\bm s}\|<\pi}\}}$ is a function that maps a rotation matrix to the corresponding rotation vector. 

        \subsubsection{\textbf{Rationality of considering noise on $\bm{\tan\theta^*}$}} In measurement model \eqref{measurement model}, we consider adding noise to $\tan\theta^*$ instead of $\theta^*$. Here, we demonstrate by simulation that there is a negligible difference between these two noise-adding mechanisms. We vary the noise intensities and compare the RMSEs of the two mechanisms. The result is shown in Fig.~\ref{Noise_injection_way}. Compared with adding noise to $\tan \theta^*$, adding noise to $\theta^*$ has an identical precision on the translation. This is because the estimation of the translation solely depends on distance measurements. As for the rotation estimation, adding noise to $\tan \theta^*$ exhibits a negligible increase in RMSE. The similar performance suggests that adding noise to $\tan \theta^*$ is reasonable and does not compromise  estimation accuracy.

        \subsubsection{\textbf{Rationality of using a single GN iteration}}
            We claim that given a $\sqrt{n}$-consistent initial estimator, a single GN iteration is enough to reach the minimum variance in the large-sample regime. To substantiate this claim, we set the noise intensity to $\sigma_{d} = 10^{-3}$ m and $\sigma_{\theta} = 10^{-3}$, and compare the RMSEs for varied GN iteration numbers. As demonstrated in Fig.~\ref{GN_compare}, when the number of points is large, the RMSE from a single GN iteration aligns with that from multiple iterations, achieving the CRLB. Even when the point number is small,  the estimation accuracy improvement with more GN iterations is marginal. 

		\subsubsection{\textbf{Influence of noise intensity}} We set the point number as $n=14$ and let both standard deviation $\sigma_d$ (m) and $\sigma_{\theta}$ vary from 
        $10^{-5}$ to $10^{-1}$. The result is presented in Fig.~\ref{RMSE_varied_noise}. When the noise intensity is small, the BESTAnP, Non-app, and Combined+CIO algorithms accurately estimate the sonar's pose, while the App algorithm exhibits a constant large RMSE.  
        As noise intensity increases, the Non-app algorithm becomes the worst one, while the App method performs even better than the CRLB. This may be because the App algorithm imposes a strong priori that the elevation angle of each point is 0, which restrains the effect of noise. The RMSEs of the Combined+CIO and BESTAnP algorithms are comparable when the noise intensity is large.
        
        

        \subsubsection{\textbf{Influence of point number}} We fix $\sigma_{d}=10^{-3}$ m and $\sigma_{\theta}=10^{-3}$. The number of points is set to be 10, 30, 90, 270, and 1,000, respectively. The RMSE comparison is presented in Fig.~\ref{RMSE_varied_pt_num}. As observed, the RMSE of BESTAnP decreases linearly with respect to the point number in the double-log axes, which demonstrates its $\sqrt{n}$-consistent property. Moreover, it asymptotically achieves the CRLB as the point number increases. 
        The App and Non-app algorithms are inferior to BESTAnP and cannot reach the CRLB. The Combined+CIO algorithm outperforms the CRLB at some point numbers. This is because, in its iterative optimization, it additionally considers the constraint that each point is within the vertical aperture of the sonar. 
        We also present the average running time of each algorithm in TABLE~\ref{Running time}. The adopted platform is Raspberry Pi 5 Model B with CPU Cortex-A76 running Debian GNU/Linux 12.
         The proposed BESTAnP algorithm is extremely efficient, over ten times faster than the Combined+CIO method. The computational advantage of our algorithm is due to its six-degree closed-form solution and only a single GN refinement. Note that the frame frequency of an FLS is typically tens of Hz. Hence, our BESTAnP features real-time capacity in resource-constrained platforms, while the compared algorithms do not.


        \subsubsection{\textbf{Influence of point distribution}} The noise intensity is fixed at $\sigma_{d}=10^{-3}$ m and $\sigma_{\theta}=10^{-3}$. The ranges of the azimuth $\theta$ and elevation $\phi$ are set as $[-3 \alpha^{\circ},3 \alpha^{\circ}]$ and $[-\alpha^{\circ},\alpha^{\circ}]$, where $\alpha=1, 2, \dots, 10$, respectively. For each distribution, 100 points can be seen. The result is shown in Fig.~\ref{RMSE_varied_pt_dis}. 
        When 3D points are tightly clustered, the App algorithm provides the best estimation since the small value of $\phi$ aligns with its assumption of $\cos\phi=1$. However, it becomes the worst one as the distribution disperses.
        In contrast, the BESTAnP, Non-app, and Combined+CIO algorithms exhibit decreasing RMSEs when the range of 3D points increases. This illustrates that a dispersed point distribution is generally good for AnP pose estimation.

            \begin{table}[htbp]
                \centering
                \caption{Comparison of average running time for a single estimation (ms).}
                \label{Running time}
                \begin{tabular}{c c c c c c}
                \thickhline
                Number of Points  & 10 & 30 & 90 & 270  & 1000 \\
                \hline
                BESTAnP & \textbf{1.06} & \textbf{1.35} & \textbf{2.17} & \textbf{6.22} & \textbf{34.69} \\
                App & 42.25 & 37.36 & 37.01 & 45.54 & 99.62 \\
                Non-app & 21.24 & 24.64 & 27.56 & 112.00 & 205.48 \\
                Combined+CIO & 107.84 & 113.64 & 125.96 & 227.23 & 424.05 \\
                \thickhline
                \end{tabular}
            \end{table}


        \subsection{Sonar-Based Odometry with Unknown 3D Points}\label{simulation_trajector}
        In this subsection, we integrate our BESTAnP algorithm with feature point triangulation to develop a pure sonar-based odometry system.  The 3D triangulation adopted here is ANRS~\cite{negahdaripour2018application} , which uses the distance and azimuth measurements from two sonar frames with a known relative pose to triangulate the 3D points of interest. 
       To the best of our knowledge, no study has put forward an odometry from sonar-only measurements.

        The sonar-based odometry mainly includes three components: 1) initialization; 2) AnP pose tracking; 3) 3D reconstruction. 
        At the initialization stage, the noise-corrupted poses at the first two frames are provided so that some initial 3D points are triangulated via the ANRS algorithm~\cite{negahdaripour2018application}. 
        After that, AnP and ANRS algorithms are alternately executed to form a full odometry. Specifically, the AnP algorithm is used to track new sonar frames with already reconstructed 3D points, while the ANRS algorithm triangulates new points to the map with the latest two sonar frames. To maintain robustness, we employ re-projection error as a metric to evaluate the accuracy of reconstructed points and remove potential outliers.
        The sonar's measurable distance and angle ranges are the same as those in the previous subsection.  
        The feature point density is set to ensure approximately 50 valid points are utilized by the AnP algorithm at each frame.
        The noise intensity is set as $\sigma_d=10^{-3}$ m and $\sigma_{\theta}=10^{-3}$.
        
\begin{figure}[!htbp]
\centering
 \begin{subfigure}[b]{0.54\linewidth}
      \includegraphics[width=\linewidth]{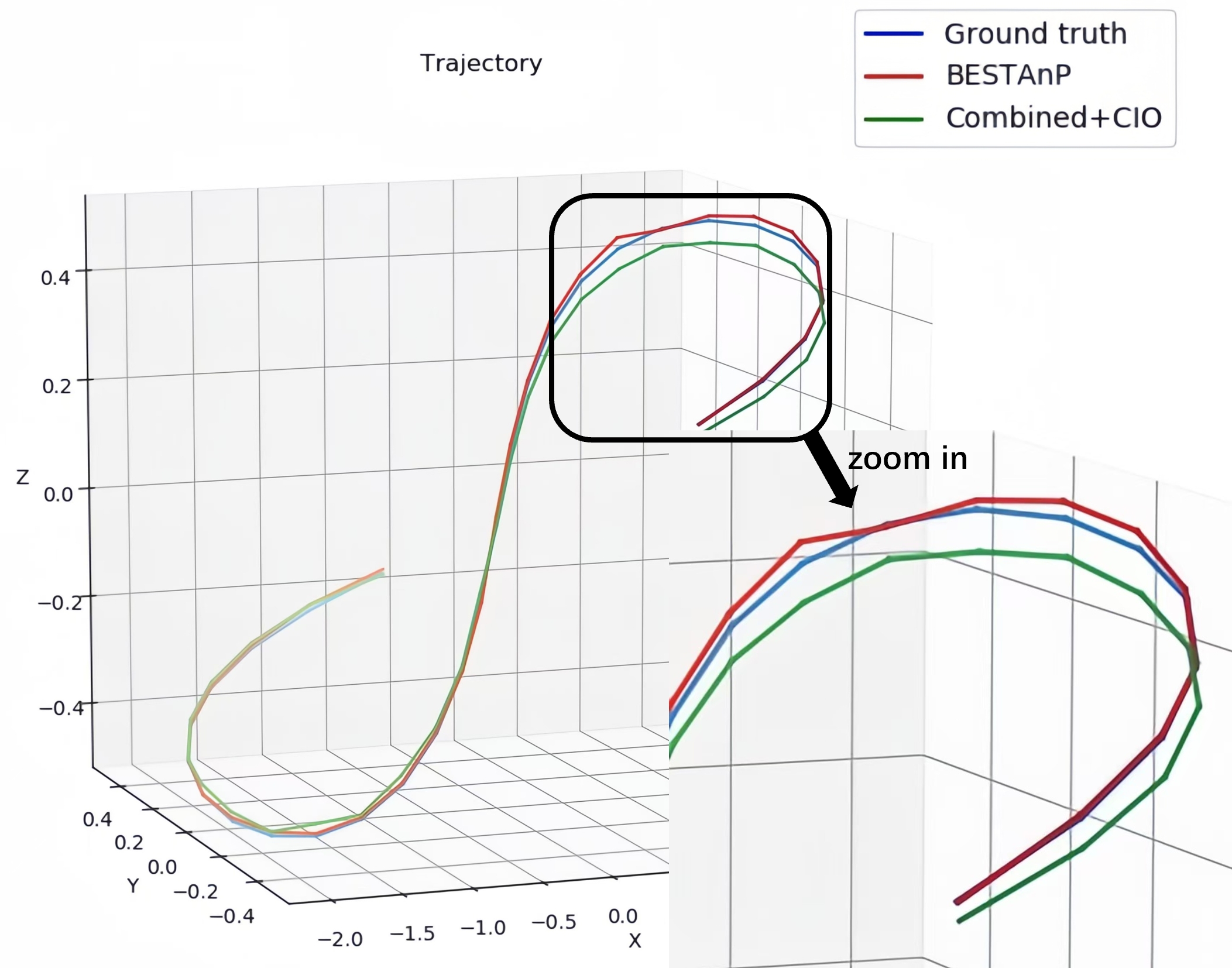}
      \caption{8-shaped}
    \label{8_odometry_simulation}
 \end{subfigure}
 \begin{subfigure}[b]{0.44\linewidth}
      \includegraphics[width=\linewidth]{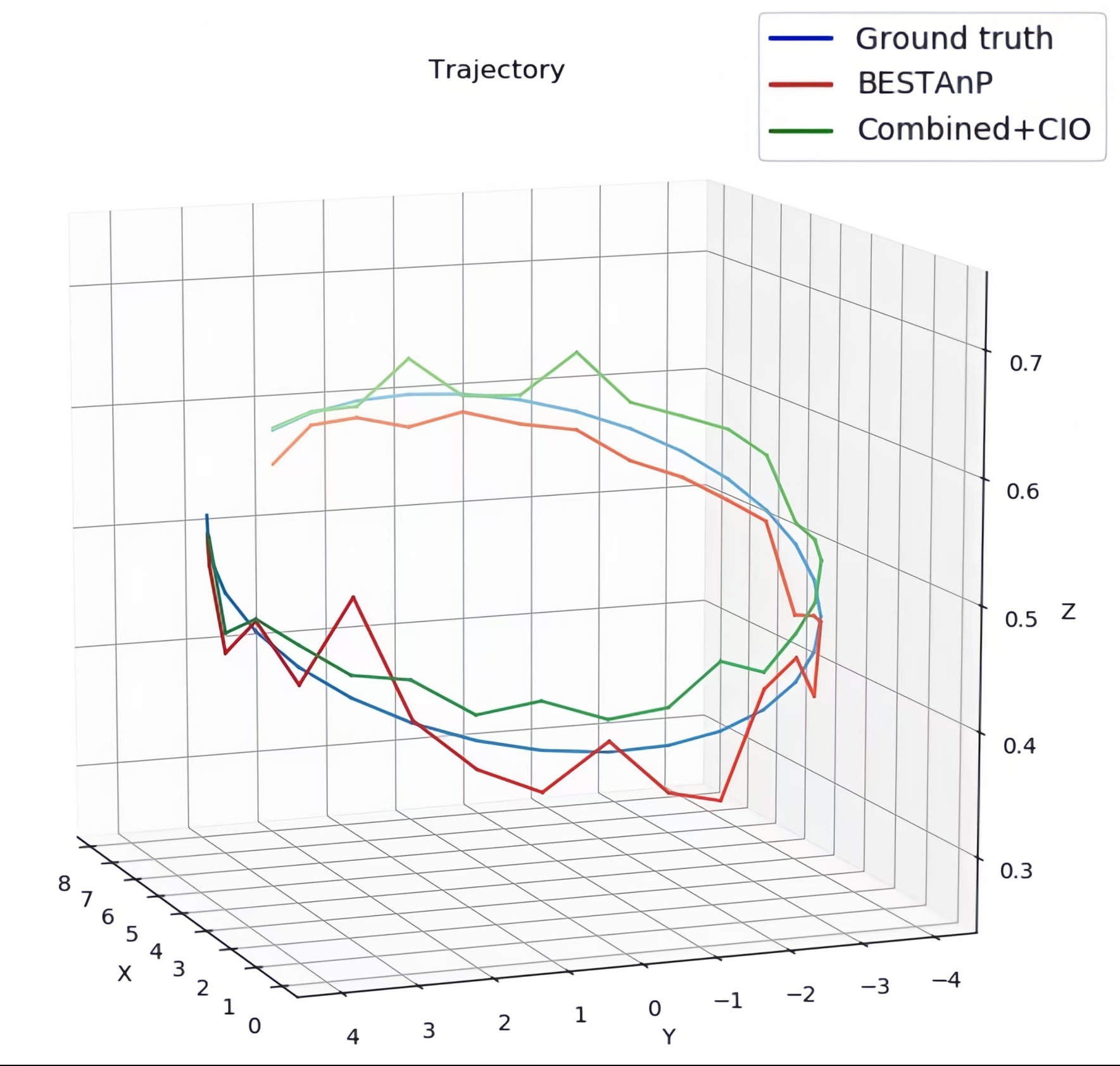}
      \caption{Circle}
    \label{9_odometry_simulation}
 \end{subfigure}
   \caption{Comparison of trajectory estimation results from BESTAnP and Combined+CIO algorithms. Color intensity indicates temporal progression, with darker shades representing more recent positions along the path.}
    \label{odometry_simulation}
\end{figure}

        We test the odometry system for two trajectories: 8-shaped and circle. 
        The absolute trajectory errors (ATE) and relative pose errors (RPE)~\cite{zhang2018tutorial} are selected as evaluation metrics and listed in TABLE~\ref{table:average_errors}. Results demonstrate that our BESTAnP algorithm gives the best pose for the 8-shaped trajectory and rotation for the circle one. The Combined+CIO method shows comparable accuracy, with best translation estimation for the circle trajectory. The Non-app method consistently exhibits the largest errors among all approaches.
        We also plot the estimated trajectories of the two best-performing methods, BESTAnP and Combined+CIO. In the 8-shaped trajectory (Fig.~\ref{8_odometry_simulation}), BESTAnP exhibits better estimation accuracy, particularly at the end stage of the trajectory where drifting errors become more apparent for the Combined+CIO method. For the circle trajectory (Fig.~\ref{9_odometry_simulation}), the two methods show similar absolute performance, though Combined+CIO maintains a slightly smoother path.


        \begin{table}[h]
        \centering
        \caption{Average errors for different trajectories and methods. The units for errors of $\bm t$ and $\bm R$ are m and $^\circ$. }
        \label{table:average_errors}
        \begin{tabular}{cccccc}
        \thickhline
        \multirow{2}{*}{\textbf{Trajectory}} & \multirow{2}{*}{\textbf{Method}} & \multicolumn{2}{c}{\textbf{ATE}} & \multicolumn{2}{c}{\textbf{RPE}} \\ \cline{3-6}
        & & $\bm t$ & $\bm R$ & $\bm t$ & $\bm R$\\
        \hline
        \multirow{4}{*}{8-shaped} 
        & BESTAnP & \textbf{0.0094} & \textbf{0.93} & \textbf{0.0088} & \textbf{0.95} \\
        & App & 0.1293 & 1.96 & 0.0884 & 2.15 \\ 
        & Non-app & 0.3618 & 5.32 & 0.3817 & 5.66 \\
        & Combined+CIO & 0.0248 & 1.54 & 0.0095 & 1.55 \\
        \hline

        \multirow{4}{*}{Circle} 
        & BESTAnP & 0.0321 & \textbf{1.14} & 0.0417 & \textbf{1.69} \\
        & App & 0.0446 & 3.29 & 0.0415 & 4.66 \\
        & Non-app & 0.6420 & 12.84  & 0.6436 & 13.82 \\
        & Combined+CIO  & \textbf{0.0303} & 1.51 & \textbf{0.0243} &  2.08 \\
        \thickhline
        
        \end{tabular}
        \end{table}

\section{Experiments on real-world data}
\begin{figure}[htbp]
			\centering
   \begin{subfigure}[b]{0.31\textwidth}
				\centering
				\includegraphics[width=1\textwidth]{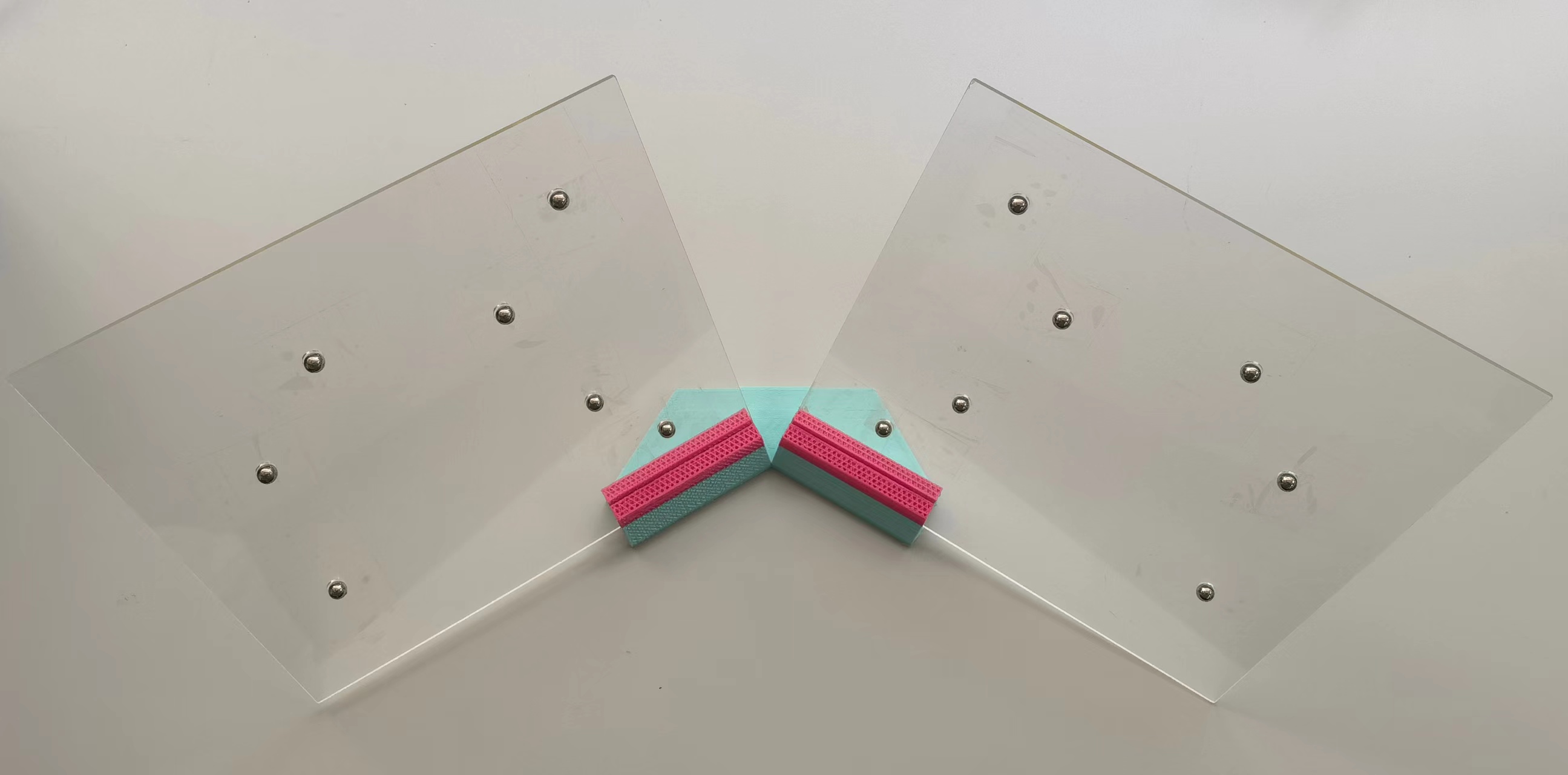}
				\caption{Dual-palne 3D structure}
				\label{Dual-palne 3D structure}
			\end{subfigure}
\begin{subfigure}[b]{0.16\textwidth}
				\centering
				\includegraphics[width=1\textwidth]{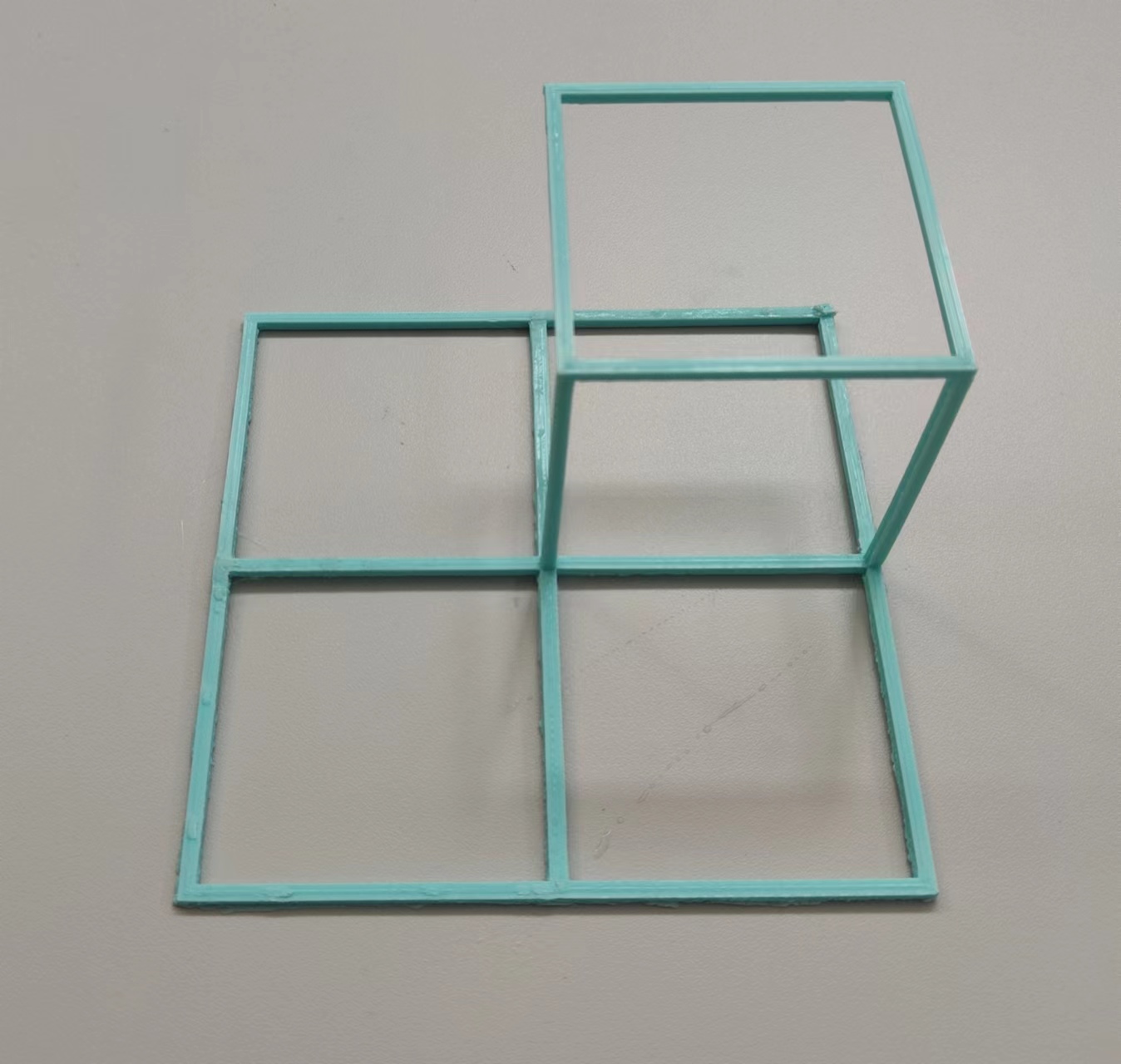}
				\caption{Cube 3D structure}
				\label{Cube 3D structure}
			\end{subfigure}
   \begin{subfigure}[b]{0.47\textwidth}
				\centering
				\includegraphics[width=1\textwidth]{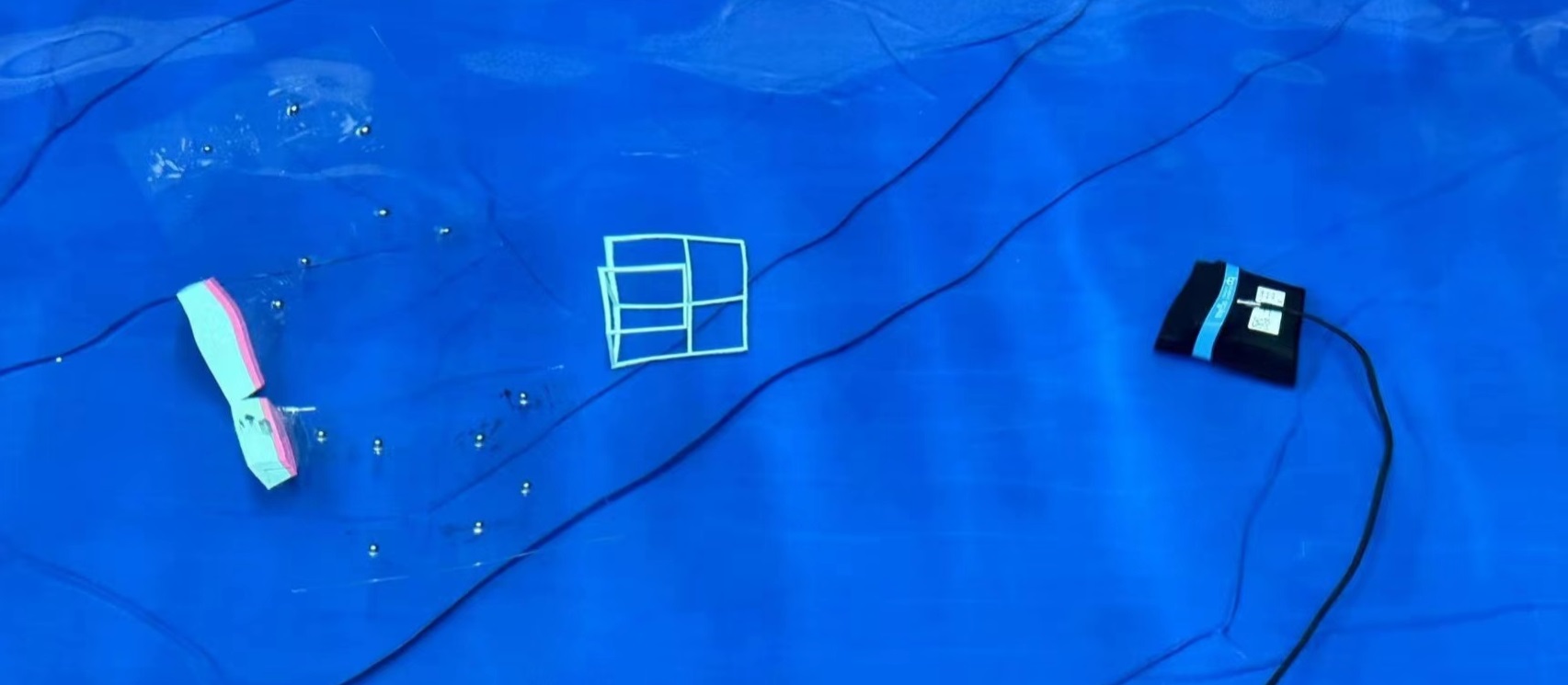}
				\caption{Experiment scenario}
				\label{Real-world experimental environment}
			\end{subfigure}
			\caption{Laboratory pool experiment.}
			\label{Laboratory pool experiment}
		\end{figure}

In this section, we demonstrate the practical utility of our algorithm through real-world experiments. The experiments are conducted in a 3m$\times$4m$\times$1.5m laboratory pool. We construct two 3D structures: the dual-plane structure shown in Fig.~\ref{Dual-palne 3D structure} consists of two non-coplanar acrylic plates and 14 steel balls scattered on them; the cube structure shown in Fig.~\ref{Cube 3D structure} is constructed by 3D printing, featuring a cube stacked on top of a quartering square. The steel balls on the dual-pane structure and corner points of the cube structure are used as feature points for sonar pose estimation. We use the Oculus 1200d sonar operating in the 2.1MHz mode with a 60$^{\circ}$ horizontal aperture and a 12$^{\circ}$ vertical aperture. The experiment scenario is shown in Fig.~\ref{Real-world experimental environment}. 

\subsection{Gaussian Assumption Verification}
\begin{figure}[htbp]
\begin{subfigure}[b]{0.22\textwidth}
				\includegraphics[width=1\textwidth]{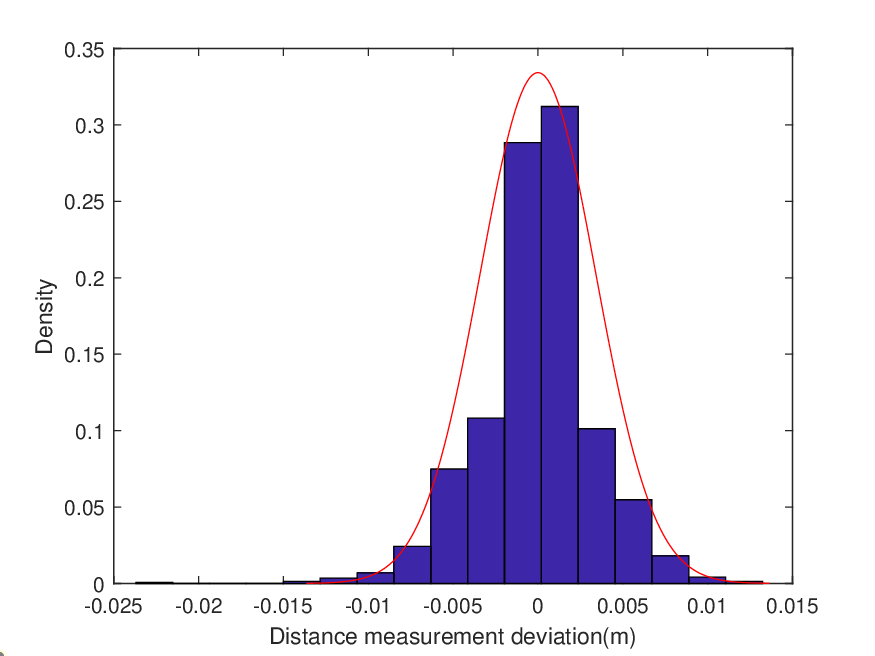}
				\caption{Distance deviations}
				\label{distance_error}
			\end{subfigure}
   \begin{subfigure}[b]{0.22\textwidth}
				\includegraphics[width=1\textwidth]{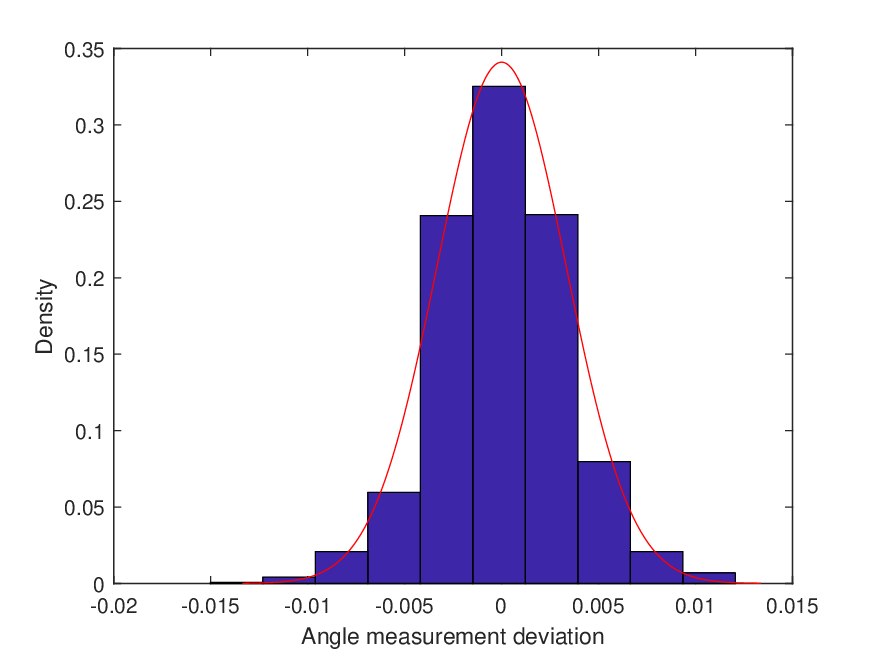}
				\caption{Azimuth deviations}
				\label{theta_error}
			\end{subfigure}
			\caption{Empirical distributions of sonar measurement deviations.}
			\label{error}
		\end{figure}

We set the sonar and the dual-plane structure at fixed positions and obtained 100 repeated measurements for each steel ball. Denote the $j$-th measurement for the $i$-th ball as $d_{ij}$ and $\tan \theta_{ij}$. The measurement sample mean for each ball is $\bar d_i=\sum_{j=1}^{100} d_{ij}/100$ and $\tan \bar \theta_i=\sum_{j=1}^{100}\tan \theta_{ij}/100$. Then, the deviation of each measurement from the sample mean can be calculated as $\Delta d_{ij}=d_{ij}-\bar d_i$ and $\Delta \tan \theta_{ij}=\tan \theta_{ij}-\tan \bar \theta_i$. We draw the empirical distributions of $\Delta d_{ij}$ and $\Delta \tan \theta_{ij}$ in Fig.~\ref{error}. 
The plot shows that both empirical distributions coincide well with a Gaussian distribution, verifying the rationality of our assumption of Gaussian measurement noises. 
The sample standard deviation of distance measurements is 0.0034 m and that of the tangent of angle measurements is 0.0033.

        \subsection{Absolute Pose Estimation with Known 3D Points}



        We first fix the dual-plane structure in the pool and move the sonar to scan it at several random poses. These poses with respect to the dual-plane structure are then estimated using feature point measurements. 
        The same test is then conducted for the cube structure. Since the models of the two structures are known, the 3D coordinates of the feature points in their corresponding structure frame are available. For data association, the correspondences between 3D points and 2D measurements are established manually. 
        After obtaining the estimated sonar pose, we reproject the 3D points onto the sonar image using the estimated pose.
        Average reprojection error (ARE) is utilized for evaluation:
        \begin{equation*}
            \mathrm{ARE} \!= \!{\frac{1}{N}\sqrt{\sum_{i=1}^N\|^{SI} \!\hat{\bm{q}}_i-^{SI} \!\bm{q}_i\|^2}},
        \end{equation*}
        where $N$ is the number of points, $^{SI} \!\bm{q}_i$ is the measurement, and $^{SI} \!\hat{\bm{q}}_i$ is the reprojection point using the estimated pose.

          \begin{table}[ht]
        \small
            \centering
            \caption{ARE and running time in the real-world experiment. The units for ARE and running time are mm and ms.}
            \label{Reprojection_errors}
            
            \begin{tabular}{c c  c c}
            \thickhline
             \textbf{Structure} & \textbf{Method}  & \textbf{ARE} & \textbf{Running time}\\
            \hline
            \multirow{4}{*}{Dual-plane} & BESTAnP & 1.774 & \textbf{0.89}\\ 
            & App  & 1.597 & 2.45\\
             & Non-app  & 22.109 & 9.32\\
            & Combined+CIO  &\textbf{1.112} & 28.77\\
            \hline
            \multirow{4}{*}{Cube} & BESTAnP  & 0.876 & \textbf{0.86}\\ 
            & App  &	1.356 &2.44\\
            & Non-app &	5.997 & 5.30\\
            & Combined+CIO  & \textbf{0.773} & 24.88\\
            \thickhline
            \end{tabular}
        \end{table}

        The ARE and running time  are listed in TABLE~\ref{Reprojection_errors}. The Combined+CIO algorithm has the smallest reprojection error for both structures, while the Non-app method performs worst. Our BESTAnP algorithm is slightly inferior to the App method for the dual-plane but performs better for the cube. This is because the dual-plane structure has a narrower elevation angle range, which aligns more with the assumption of $\cos \phi=1$ in the App algorithm.
        Regarding time complexity, maybe the initial value is better and CIO needs fewer iterations, the compared three algorithms cost less time than in the previous simulation. Nonetheless, our BESTAnP is much more efficient than them, costing approximately only 3\% of that of the Combined+CIO algorithm. 

       \subsection{Sonar-Based Odometry with Unknown 3D Points}
       We regard the 14 balls on the dual-plane structure as unknown 3D points and triangulate them with five provided poses in the initialization stage. The five initial poses can be given by another odometry pipeline, for example, the preintegration of an inertial measurement unit (IMU). Here, we simulate the preintegration of IMU by adding noises to the five poses given by an underwater motion capture system. The triangulation method is similar to the ANRS algorithm used in Section~\ref{simulation_trajector}.  
       We run two trajectories: an 8-shaped trajectory and a LIAS-shaped trajectory. Along both paths, the 14 balls on the dual-plane structure are always within the FOV of the sonar so that the whole trajectories can be estimated by an AnP algorithm.
       The ground truth trajectories are provided by the underwater motion capture system. 
       
\begin{figure}[!htbp]
\centering
 \begin{subfigure}[b]{0.49\linewidth}
      \includegraphics[width=\linewidth]{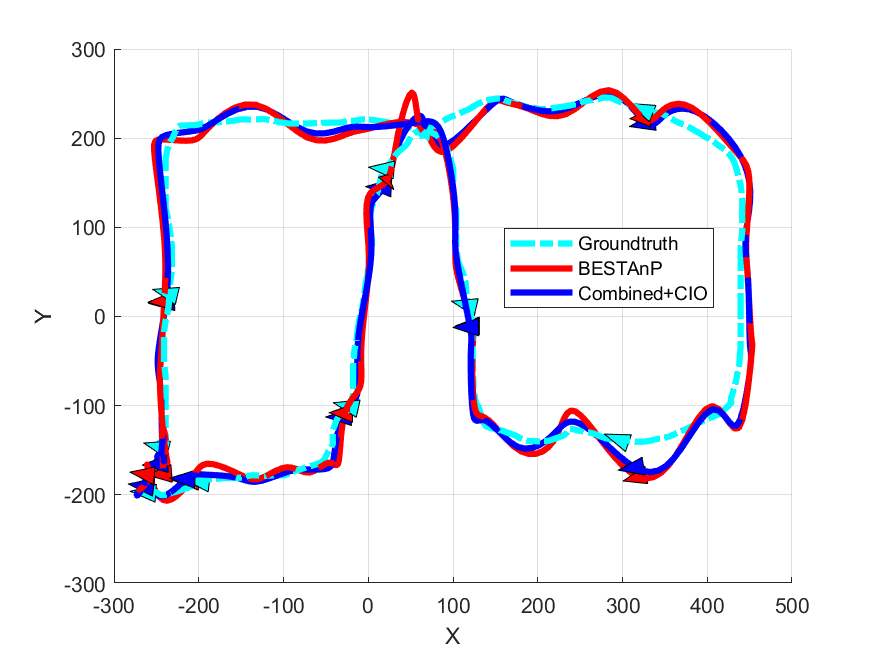}
      \caption{8-shaped}
    \label{8_odometry_real_world}
 \end{subfigure}
 \begin{subfigure}[b]{0.49\linewidth}
      \includegraphics[width=\linewidth]{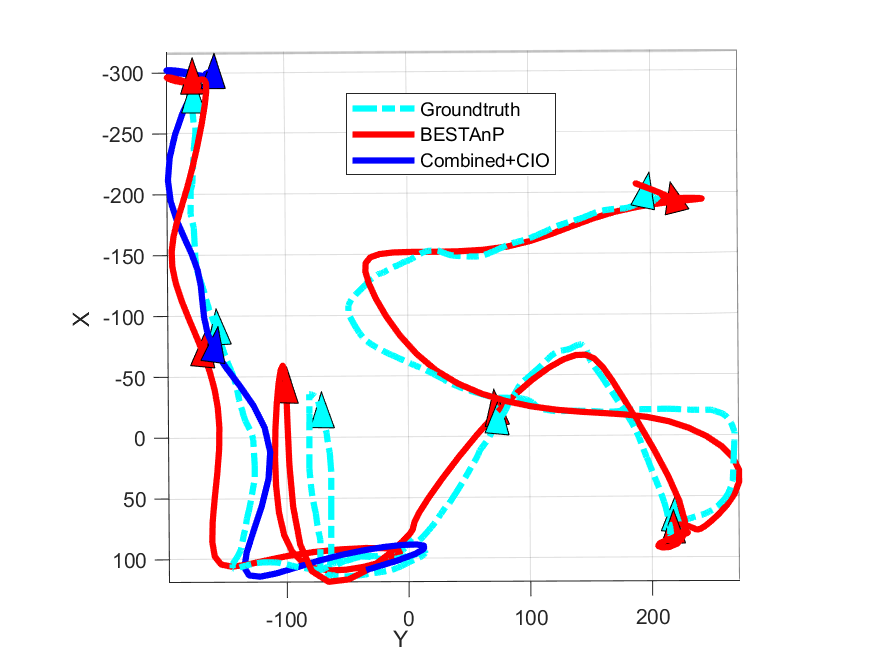}
      \caption{LIAS-shaped}
    \label{LIAS_odometry_real_world}
 \end{subfigure}
   \caption{Trajectory estimation in real-world experiments.}
    \label{odometry_real_world}
\end{figure}

       We calculate ATE and RPE for performance analysis. The trajectory estimation results are shown in Fig.~\ref{odometry_real_world} and the trajectory errors are listed in TABLE~\ref{Real_world_trajectory_error}. For the compared algorithms, we only present the results of the best one (Combined+CIO).
       In visualizing the estimated trajectories, we terminate plotting a trajectory when it first comes across an abnormal estimation---an estimation that is far away from the groundtruth.
       We see that our algorithm succeeds in plotting the whole trajectory close to the groundtruth in both cases. Although the Combined+CIO algorithm achieves slightly better accuracy than ours for the 8-shaped trajectory, it only completes a partial estimation for the LIAS-shaped path. This is because its CIO procedure is sensitive to the initial value and not so stable. The unsatisfactory performance of the Combined+CIO algorithm on translation estimation for the LIAS-shaped trajectory is also reflected in TABLE~\ref{Real_world_trajectory_error}.  
        \begin{table}[h]
            \centering
       \caption{Average errors for different trajectories and methods. The units for errors of $\bm t$ and $\bm R$ are mm and $^\circ$.}
            \label{Real_world_trajectory_error}
            \begin{tabular}{cccccc}
           \thickhline
             \multirow{2}{*}{\textbf{Trajectory} } & \multirow{2}{*}{\textbf{Method}} & \multicolumn{2}{c}{\textbf{ATE}}   & \multicolumn{2}{c}{\textbf{RPE}}\\\cline{3-6}
              & & $\bm t$ & $\bm R$  & $\bm t$ & $\bm R$\\
            \hline
            
            \multirow{2}{*}{8-shaped} & BESTAnP & 55.76  & 0.98& 64.94 & 0.89\\ 
            & Combined+CIO & \textbf{51.14} & \textbf{0.58} & \textbf{51.35} & \textbf{0.69} \\
            \hline

            \multirow{2}{*}{LIAS-shaped} & BESTAnP & \textbf{44.33}  & \textbf{0.51} & \textbf{34.67} & \textbf{0.23}\\ 
            & Combined+CIO & 584.46  & 0.69 & 816.56& 0.92\\
           \thickhline
            \end{tabular}
        \end{table}

\section{Conclusion and future work}
    In this letter, we explored the intrinsic relationship between the sonar's distance measurements and its position, as well as the relationship between the azimuth angle measurements and its orientation. Based on these connections, we proposed a bi-step AnP algorithm (BESTAnP) that provides a closed-form solution for the six-degree sonar pose. A single GN iteration is further utilized to refine the initial consistent estimator. Experimental results showed that BESTAnP is efficient, accurate, and stable. 
    For future work, how to enhance the applicability of the AnP algorithm in unknown environments is of great value. In particular, it is interesting to design a feature point association method between sonar images, tailored to supply abundant correspondences so that our consistent AnP algorithm can fully leverage its advantage.

        \appendices

\section{Proof of Lemma~\ref{angle_noise_estimation}} \label{proof_angle_noise}
    The proof is mainly based on the following lemma:
    \begin{lemma}[{\cite[Lemma 1]{zeng2024optimal}}] \label{lemma_largest_eig}
	Let ${\bf R}$ and ${\bf S}$ be two real symmetric matrices and ${\bf Q}={\bf R}+{\bf S}$. If ${\bf Q}$ is positive-definite and ${\bf R}$ is positive-semidefinite with $0$ eigenvalues, then $\lambda_{\rm max}( {\bf Q}^{-1} {\bf S})=1$.
\end{lemma}
From Lemma~\ref{Lemma_1}, we have
\begin{equation*}
    {\bm Q}={\bm Q}^*+\sigma_{\theta}^2 {\bm S} + O_p(1/\sqrt{n}).
\end{equation*}
This implies that $\lim_{n \rightarrow \infty} {\bm Q}={\bm Q}^*+\sigma_{\theta}^2 {\bm S}$. On the one hand, in the noise-free case, ${\bm B}^* {\bm r}^*=0$, which implies that ${\bm B}^*$ does not have full column rank. Hence, ${\bm Q}^*$ is a positive-semidefinite matrix with $0$ eigenvalues. 
On the other hand, since ${\bm B}$ is corrupted by random noises, given $n \geq 6$ point correspondences, ${\bm B}$ has full column rank with probability one, i.e., ${\bm Q}$ is almost surely positive-definite. Then, according to Lemma~\ref{lemma_largest_eig}, we have $\lim_{n \rightarrow \infty} 1/\lambda_{\rm max}( {\bm Q}^{-1} {\bm S})=\sigma_{\theta}^2$, i.e., $\hat \sigma_{\theta}^2$ converges to $\sigma_{\theta}^2$. Moreover, since ${\bm Q}-\sigma_{\theta}^2 {\bm S}$ converges to ${\bm Q}^*$ with a rate of $1/\sqrt{n}$, the convergence rate of $\hat \sigma_{\theta}^2$ is also $1/\sqrt{n}$, which completes the proof.
  
        \section{Proof of theorem~\ref{theorem_Q}} \label{appendix_consistent_R}
            The following lemma plays a vital role in this proof:
            \begin{lemma}{(\cite{9855392}, Lemma 4):}\label{Lemma_1}
            Let $\{X_i\}$ be a sequence of independent random variables with $\mathbb{E}[X_i]=0$ and $\mathbb{E}[X_i^2] \leq \phi \leq \infty$ for all $i$. Then, there holds $\sum_{i=1}^n X_i/n = O_p(1/\sqrt{n})$.
		\end{lemma}
  
            Let $\bm W_i^* =  (^W \! \bm p_i - {\bm t}^*) \bm (^W \! \bm p_i - {\bm t}^*)^
            \top$, The noise-free matrix $\bm Q^*$ is calculated as follows:

            \begin{eqnarray*}
			{\bm Q^*} = {1 \over n}
			\left[\begin{array}{cc}
				\sum_{i=1}^{n} \tan^2\theta_i^* \bm W_i^* &  -\sum_{i=1}^{n} \tan\theta_i^* \bm W_i^* \\
				-\sum_{i=1}^{n} \tan\theta_i^* \bm W_i^* & \sum_{i=1}^{n} \bm W_i^*
			\end{array}\right].
		  \end{eqnarray*}

            Under the influence of measurement noise, i.e., $\tan\theta_i = \tan\theta_i^*+\epsilon_{\theta_i}$, and the $\sqrt{n}$-consistent estimator $\hat{\bm t}^{BE}$ for translation, the noisy matrix $\bm Q$ is obtained as follows:
        \begin{footnotesize}
            \begin{align*}
            {\bm Q} &=  {1 \over n}
			\left[\begin{array}{cc}
				\sum_{i=1}^{n} (\tan\theta_i^*+\epsilon_{\theta_i})^2 \bm W_i &  -\sum_{i=1}^{n} (\tan\theta_i^*+\epsilon_{\theta_i}) \bm W_i \\
				-\sum_{i=1}^{n} (\tan\theta_i^*+\epsilon_{\theta_i})\bm W_i & \sum_{i=1}^{n} \bm W_i
			\end{array}\right] \\
		&= \bm Q^* +  
		\left[\begin{array}{cc}
			{1 \over n}\sum_{i=1}^{n} (2\tan\theta_i^*\epsilon_{\theta_i}+\epsilon_{\theta_i}^2) \bm W_i &  -{1 \over n}\sum_{i=1}^{n} \epsilon_{\theta_i} \bm W_i \\
			-{1 \over n}\sum_{i=1}^{n} \epsilon_{\theta_i} \bm W_i & \bm 0_{3 \times 3}
		\end{array}\right]\\
         &~~~~+O_p({1 \over \sqrt{n}}) \\
		&= \bm Q^* + 
		\left[\begin{array}{cc}
			\hat \sigma_{\theta}^2 {1\over n}\sum_{i=1}^{n} \bm W_i &  \bm 0_{3 \times 3} \\
			\bm 0_{3 \times 3} & \bm 0_{3 \times 3}
		\end{array}\right] + O_p({1 \over \sqrt{n}}),
    \end{align*}
        \end{footnotesize}
        where the second ``$=$'' holds since $\hat{\bm t}^{BE}$ is a $\sqrt{n}$-consistent estimator of $\bm t^*$ (Theorem~\ref{theorem_t}), and the third ``$=$'' holds according to Lemmas~\ref{angle_noise_estimation} and~\ref{Lemma_1}. Therefore, $\bm Q^{\rm BE} = \bm Q -\bm C=\bm Q^* + O_p(1/\sqrt{n})$, i.e., $\bm Q^{\rm BE}$ is a $\sqrt{n}$-consistent estimator of $\bm Q^*$. In addition, since the eigendecomposition is a continuous function that can maintain the $\sqrt{n}$ consistency. Hence, $\sqrt{2}{\bm v}_{\lambda_{\rm min}}^\mathrm{BE}$ is a $\sqrt{n}$-consistent estimator of $\bm r^*$ up to a sign. Finally, $\hat{\bm r}^\mathrm{BE}$ corrects the sign of $\sqrt{2}{\bm v}_{\lambda_{\rm min}}^\mathrm{BE}$, which completes the proof.

\section{The formulas for the GN iteration} \label{formula_GN}

Note that the rotation matrix must be optimized within the constraints of ${\rm SO}(3)$. Thus, we cannot directly calculate the Jacobian matrix by taking the derivative with respect to $\bm R$. Instead, we optimize over its tangent space. Given a rotation matrix $\hat{\bm R}^\mathrm{BE} \in {\rm SO}(3)$ and any vector ${\bm s}=[s_1 ~ s_2 ~ s_3]^\top \in \mathbb R^3$, the matrix $\hat{\bm R}^\mathrm{BE}\exp({\bm s}^{\wedge})$ is also a rotation matrix~\cite{zeng2023cpnp}, where ${\rm exp}$ denotes the matrix exponential and the ``hat'' operator generates a skew-symmetric matrix as follows:
         \begin{align*}
            \bm{s}^{\wedge}=\left[\begin{array}{ccc}
            0 & -s_3 & s_2 \\
            s_3 & 0 & -s_1 \\
            -s_2 & s_1 & 0
            \end{array}\right].
        \end{align*}
        Hence, we can refine the rotation matrix by optimizing the vector $\bm s$.

        Define $\bm R = \hat{\bm R}^\mathrm{BE}\exp(\bm s^{\wedge})$, then $f_{\theta_i}(\bm R,\bm t)$ can be transformed into a function of $\bm s$:
        \begin{align}
            f_{\theta_i}(\bm s,\bm t) = \tan\theta_i-\frac{\bm e_2^\top (\bm \hat{\bm R}^\mathrm{BE}\exp(\bm s^{\wedge}))^\top({^W \! \bm p_i} - \bm t)}{\bm e_1^\top (\bm \hat{\bm R}^\mathrm{BE}\exp(\bm s^{\wedge}))^\top(^W \! \bm p_i - \bm t)}.
        \end{align}
        The GN iteration starts with $\bm s = \bm 0_{3\times 1}$. 
        Define $\bm u_i = {^W \! \bm p_i} - \bm t$ and 
		\begin{align*}
			\Psi &= \left.{\partial  \text{vec}(\exp(s^{\wedge})) \over \partial \bm s^\top}\right|_{\bm s=\bm 0},\\
			g_i( \bm s, \bm t) &= \bm e_2^\top (\bm \hat{\bm R}^\mathrm{BE}\exp(\bm s^{\wedge}))^\top {\bm u_i}, \\
			h_i(\bm s, \bm t) &= \bm e_1^\top (\bm \hat{\bm R}^\mathrm{BE}\exp(\bm s^{\wedge}))^\top {\bm u_i},
		\end{align*}
        where ${\rm vec}$ denotes the vectorization operation. The partial derivatives of the functions $f_{d_i}({\bm t})$ and $f_{\theta_i}({\bm s},{\bm t}) $ with respect to $\bm s$ and $\bm t$ can be calculated as 
        \begin{align*}
            \left.{\partial f_{d_i}({\bm t}) \over \partial \bm s^\top}\right|_{\bm s=\bm 0} &= [0~ 0~ 0],  ~~~~
		  \left.{\partial f_{d_i}({\bm t}) \over \partial \bm     t^\top}\right|_{\bm s=\bm 0} = {\bm u_i^\top \over \Vert \bm   u_i\Vert}, \\
            \left.{\partial f_{\theta_i}({\bm s},{\bm t}) \over \partial \bm s^\top}\right|_{\bm s=\bm 0} &= \!{(g_i(\bm s, \bm t)\bm e_1^\top \!-\! h_i(\bm s, \bm t)\bm e_2^\top)\otimes (\bm u_i^\top  \hat{\bm R}^{\mathrm{BE}}) \Psi \over h_i(\bm s, \bm t)^2}, \\
            \left.{\partial f_{\theta_i}({\bm s},{\bm t}) \over \partial \bm t^\top}\right|_{\bm s=\bm 0} &= {(h_i( \bm s, \bm t)\bm e_2^\top - g_i( \bm s, \bm t)\bm e_1^\top) \hat{\bm R}^{\mathrm{BE}\top} \over  h_i( \bm s, \bm t)^2},
        \end{align*}
        where $\otimes$ denotes the Kronecker product.

        Then, we can obtain the following residual vector and Jacobian matrix:
        \begin{align*}
                \bm {\Delta} \! =\! \left[\begin{array}{c}
				f_{d_1}(\hat{\bm t}^\mathrm{BE})/\hat \sigma_d \\
				f_{\theta_1}(\bm 0,\hat{\bm t}^\mathrm{BE})/\hat \sigma_\theta \\
				\vdots \\
				f_{d_n}(\hat{\bm t}^\mathrm{BE})/\hat \sigma_d\\
				f_{\theta_n}(\bm 0,\hat{\bm t}^\mathrm{BE})/\hat \sigma_\theta  \\
			\end{array}\right]\!\!, 
			\bm J \!= \!\!\left.\left[\begin{array}{cc}
				{\partial f_{d_1}({\bm t})/\hat \sigma_d \over \partial [\bm s^\top~ \bm t^\top]}  \\
				{\partial f_{\theta_1}({\bm s},{\bm t})/\hat \sigma_\theta \over \partial [\bm s^\top~ \bm t^\top]}\\
				\vdots \\
				{\partial f_{d_n}({\bm t})/\hat \sigma_d \over \partial [\bm s^\top~ \bm t^\top]}  \\
				{\partial f_{\theta_n}({\bm s},{\bm t})/\hat \sigma_\theta \over \partial [\bm s^\top~ \bm t^\top]} \\
			\end{array}\right]\right|_{\bm s=\bm 0,{\bm t}=\hat{\bm t}^\mathrm{BE}}.
		\end{align*}
        The single GN iteration is given by
        \begin{equation} \label{GN}
        \begin{aligned}
        \!\left[\!\!\begin{array}{c}
        \hat{\bm{s}}^{\mathrm{GN}} \\
        \hat{\bm{t}}^{\mathrm{GN}}
        \end{array}\!\!\right]& \!=\!\left[\!\! \begin{array}{c}
        \bm{0} \\
        \hat{\bm{t}}^{\mathrm{BE}}
        \end{array}\!\! \right] \!\!-\!\!  \left(\bm{J}^{\top}  \bm{J} \right)^{-1}\!\!\!\bm{J}^{\top}  \bm \!\Delta, \\
        \hat{\bm R}^\mathrm{GN} &\!= \hat{\bm R}^\mathrm{BE}\exp\left( \hat{\bm{s}}^{\mathrm{GN}^\wedge} \right).
        \end{aligned}
        \end{equation}
    
	\section*{acknowledgement}
	The authors would like to thank Shanghai ChingMU Vision Technology for providing us with an  underwater motion capture system for real-world experiments.
	

 \ifCLASSOPTIONcaptionsoff
  \newpage
\fi

\small
\bibliographystyle{IEEEtran}
\bibliography{bibfile}




\end{document}